\documentclass[pdflatex,sn-mathphys-num]{sn-jnl}

\usepackage{graphicx}%
\usepackage{multirow}%
\usepackage{amsmath,amssymb,amsfonts}%
\usepackage{amsthm}%
\usepackage{mathrsfs}%
\usepackage[title]{appendix}%
\usepackage{xcolor}%
\usepackage{textcomp}%
\usepackage{manyfoot}%
\usepackage{booktabs}%
\usepackage{algorithm}%
\usepackage{algorithmicx}%
\usepackage{algpseudocode}%
\usepackage{listings}%
\usepackage{url}

\theoremstyle{thmstyleone}%
\theoremstyle{thmstyletwo}%

\theoremstyle{thmstylethree}%

\begin{document}

\title[Article Title]{Plug-and-play Class-aware Knowledge Injection for Prompt Learning with Visual-Language Model}


\author[1]{\fnm{Junhui} \sur{Yin}}

\author*[2]{\fnm{Nan} \sur{Pu}}\email{n.pu@outlook.com}

\author[3]{\fnm{Xinyu} \sur{Zhang}}

\author[4]{\fnm{Lingfeng} \sur{Yang}}

\author[5]{\fnm{Lin} \sur{Wu}}

\author[6]{\fnm{Xiaojie} \sur{Wang}}

\author*[2]{\fnm{Zhun} \sur{Zhong}}\email{zhunzhong007@gmail.com}

\affil[1]{\orgname{University of Science and Technology Beijing},  \country{China}}

\affil[2]{\orgname{Hefei University of Technology}, \country{China}}

\affil[3]{\orgname{University of Auckland}, \country{New Zealand}}

\affil[4]{\orgname{Nanjing University of Science and Technology}, \country{China}}

\affil[5]{\orgname{Swansea University}, \country{U.K.}}

\affil[6]{\orgname{Beijing University of Posts and Telecommunications},  \country{China}}

\abstract{
Prompt learning has become an effective and widely used technique in enhancing vision-language models (VLMs) such as CLIP for various downstream tasks, particularly in zero-shot classification within specific domains. Existing methods typically focus on either learning class-shared prompts for a given domain or generating instance-specific prompts through conditional prompt learning. While these methods have achieved promising performance, they often overlook class-specific knowledge in prompt design, leading to suboptimal outcomes. The underlying reasons are: 1) class-specific prompts offer more fine-grained supervision compared to coarse class-shared prompts, which helps prevent misclassification of data from different classes into a single class; 2) compared to class-specific prompts, instance-specific prompts neglect the richer class-level information across multiple instances, potentially causing data from the same class to be divided into multiple classes. To effectively supplement the class-specific knowledge into existing methods, we propose a plug-and-play Class-Aware Knowledge Injection (CAKI) framework.
CAKI comprises two key components, \textit{i.e.}, class-specific prompt generation and query-key prompt matching. The former encodes class-specific knowledge into prompts from few-shot samples that belong to the same class and stores the learned prompts in a class-level knowledge bank. The latter provides a plug-and-play mechanism for each test instance to retrieve relevant class-level knowledge from the knowledge bank and inject such knowledge to refine model predictions. Extensive experiments demonstrate that our CAKI effectively improves the performance of existing methods on base and novel classes. Code is publicly available at \href{https://github.com/yjh576/CAKI}{this https URL}.
}


\maketitle

\section{Introduction}
\label{sec:intro}

Vision-Language foundation Models (VLMs) such as CLIP~\cite{radford2021learning} have shown powerful generalization capabilities across various downstream tasks. These VLMs are pre-trained using a task-agnostic objective designed to align web-scale image-text pairs in a shared embedding space, thereby acquiring general and transferable knowledge. This enables VLMs to encode open-vocabulary concepts, and achieve robust zero-shot prediction capabilities~\cite{jia2021scaling,lee2022uniclip}.

\par    
Although VLMs are effective in generalizing to new concepts, the massive scale of parameters and the limited availability of training data (e.g., in few-shot settings) make it impractical to fine-tune the entire model for downstream tasks. To overcome this limitation, researchers have developed parameter-efficient fine-tuning approaches~\cite{he2021towards,gao2024clip,gabeff2024wildclip} to adapt models for downstream tasks while keeping the original VLMs frozen. Among these approaches, few-shot prompt learning (FSPL)~\cite{zhou2022learning,zhou2022conditional} has become the most widely used technique to fine-tune VLMs, building on the success~\cite{prompt_nlp1,prompt_nlp2} in the Natural Language Processing community.

\begin{figure}[t]
  \centering  \includegraphics[width=0.9\textwidth]{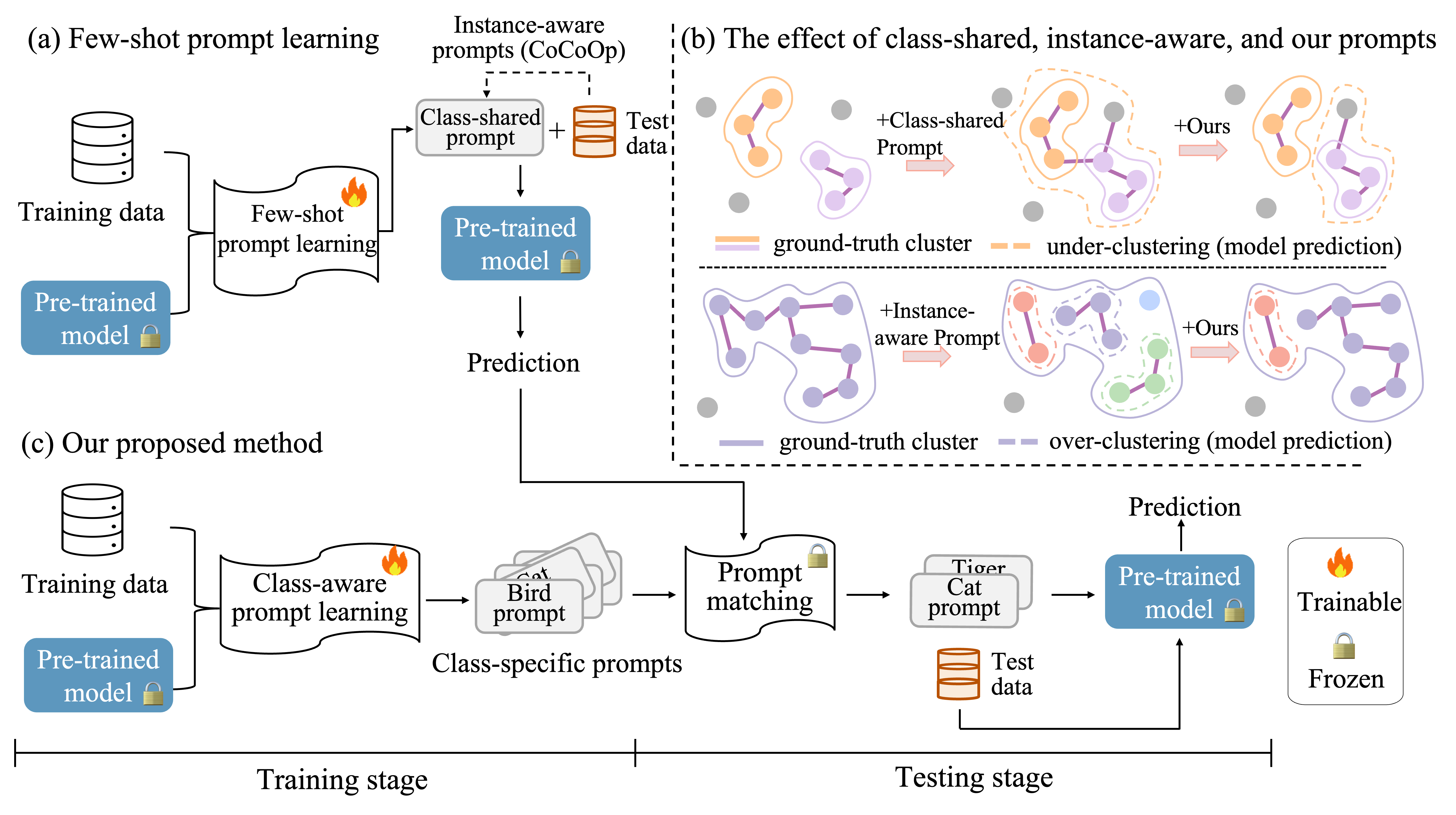}
  \caption{Comparison of our CAKI with few-shot prompt learning. (a) Existing few-shot prompt learning can learn class-shared knowledge by tuning learnable prompts on downstream tasks, which can instruct pre-trained models to predict test data. (b) The class-shared prompt exhibits excessive granularity, leading to test data from different classes being misclassified into the same class. Conversely, the instance-specific prompt overlooks class-level information, resulting in the misclassification of data from the same class into multiple classes. (c) our method encodes class-level prior knowledge into class-specific prompts and injects such knowledge into existing methods by retrieving appropriate class-specific prompts for test data.
  }\label{fig:intro}
\end{figure}

\par
To effectively adapt VLMs to downstream tasks within specific domains, existing FSPL methods either learn class-shared prompts~\cite{zhou2022learning,bahng2022exploring,jia2022visual,zang2022unified,khattak2023maple} for the specific domain or generate instance-specific prompts through conditional prompt learning~\cite{zhou2022conditional}, as shown in Figure~\ref{fig:intro} (a). However, these approaches  neglect to consider class-specific knowledge in the design of prompts. As illustrated in Figure~\ref{fig:intro} (b), this inevitably leads to suboptimal performances.
The underlying reasons are: 1) with coarse-grained class-shared prompts, applying the same prompts to the data from different classes leads to under-clustering of features. In contrast, class-specific prompts provide more fine-grained supervision, reducing the risk of misclassifying data from different classes into the same class; 2) instance-specific prompts often ignore the richer class-level information across multiple instances, resulting in the over-clustering in which data from the same class are falsely separated into multiple classes.

\par
To address these limitations, we propose a plug-and-play Class-Aware Knowledge Injection (CAKI) framework to integrate class-specific knowledge for different test instances in traditional FSPL with VLMs. Our CAKI framework mainly consists of class-specific prompt generation (CSPG) and query-key prompt matching (QKPM). CSPG aims to establish a class-level knowledge bank with key-value cache memory, where the key is the CLIP's text feature of a class name, and the value is class-specific prompts learned by the proposed class-aware prompt learning using few-shot labeled samples from the same class. Meanwhile, we employ existing FSPL approaches to learn class-shared prompts and preserve them in memory as well. 
The key issue is how to select effective class-specific prompts for different test samples. Ideally, the corresponding class prompt would be chosen for each test sample, but the labels of test samples are unknown during testing stage.
\par
To address this issue, we further propose QKPM, which leverages the fact that test samples from different classes often share significant visual similarities (e.g., cats and tigers have similar shapes and markings).
QKPM retrieves meaningful prompts from the cache memory to make more precise predictions through a coarse-to-fine process. Specifically, we regard the CLIP with class-shared prompts generated by existing FSPL methods as the coarse-grained model (CGM). For a given test sample, we use its image feature as a query and regard the prediction from CGM as the coarse-grained query-key matching scores. 
Based on the scores, we query top-K most relevant class-specific prompts from the key-value memory for the test image. These queried prompts are subsequently used to instruct the model to make multiple predictions. Considering that class-specific prompts provide varying guidance for each test sample, we further ensemble the model predictions with different prompts based on the corresponding matching scores, which can be viewed as a refinement process of coarse-grained predictions. Moreover, our QKPM is a training-free process and efficiently incorporates class-level knowledge into conventional FSPL methods for various test samples with a set of class-specific prompts, as shown in Figure~\ref{fig:intro} (c).
\par
Our contributions are summarized as follows: 1) We extend traditional FSPL to class-aware FSPL via encoding class-specific information into prompts and storing them in a key-value knowledge bank for effective retrieval and utilization.  2) To adaptively tailor class-specific prompts for each test sample, we propose a training-free query-key prompt matching method that retrieves class-relevant prompts from the knowledge bank based on query-key matching scores. 3) The proposed CAKI is a plug-and-play approach, which is compatible with most existing train-time and test-time prompt learning methods and able to further enhance the model performance.
4) Experimental results demonstrate the superior effectiveness of our CAKI over existing state-of-the-art methods on both base and novel classes.

\section{Related Work}

\textbf{Prompt learning in vision-language models.} Vision-language models like CLIP~\cite{radford2021learning} and ALIGN~\cite{jia2021scaling} are pre-trained to align the two embedding spaces on a large corpus of image-text pairs available in a contrastive self-supervised manner and have shown strong generalizability towards zero-shot recognition tasks. However, efficiently adapting them to specific downstream tasks with limited data remains a challenging problem. Recently, few-shot prompt learning (FSPL) has emerged as a new technique that achieves the desired knowledge transfer without the need for re-training of the entire model. CoOp~\cite{zhou2022learning} first proposes a FSPL framework for CLIP model by learning a set of continuous learnable vectors while keeping the CLIP parameters frozen. To enhance the model's generalization capability on out-of-distribution data, CoCoOp~\cite{zhou2022conditional} further conditions the textual prompts on visual features. Maple~\cite{khattak2023maple} proposes to learn prompts by connecting image prompts with text prompts through linear projections. PromptSRC~\cite{khattak2023self} introduces KL divergence restrictions between promptable features and vanilla features to regularize the learning of prompts for each branch. PromptKD~\cite{li2024prompt} introduces a prompt distillation framework to transfer the knowledge of a large teacher model to a small student model. Particularly, PLOT~\cite{chen2022plot}, LoCoOp~\cite{miyai2023locoop} and GalLoP~\cite{lafon2024gallop} also explores CLIP's local features for few-shot prompt learning. \textit{However, these methods either learn shared prompts for all test data or generate instance-specific prompts by conditional FSPL, which neglects class-specific knowledge in prompt design. Thus, our work aims to inject such knowledge into the existing methods in a plug-and-play manner.}

\textbf{Test-time adaptation with prompt learning.} Test-Time Adaptation (TTA)~\cite{liang2025comprehensive} has emerged as a pivotal technique for enhancing model performance on test data, without altering the training process or accessing the original training datasets. As an efficient model adaptation technique, test-time prompt tuning~\cite{shu2022test,abdul2024align} adapt VLMs to new data distribution by tuning a learnable prompt with each test sample. TPT~\cite{shu2022test} introduces a test-time prompt learning approach by learning textual prompts with an entropy minimization objective. DiffTPT~\cite{feng2023diverse} extends test-time prompt tuning by leveraging pre-trained diffusion models to
augment the diversity of test samples. 
\textit{Despite their effectiveness, the learned prompts may not sufficiently align with test data due to a lack of accurate class-specific information. Our work seeks to rectify the misalignment with learned class-specific prompts.}

\textbf{Cache memory.} 
Early works~\cite{grave2017unbounded,merity2016pointer} often store a large amount of the training dataset and aggregate information from them to boost the performance for vision or language models. However, the huge storage budget for training data hinders its applications in various real-world scenarios.  A similarity search system is proposed to alleviate this limitation by reducing the data size of cache memory. In contrast to cache memory created with data, Tip-Adapter~\cite{DBLP:journals/corr/abs-2111-03930} solves this problem by only storing few-shot visual features and one-hot labels. To address the lack of access to training data during inference, TDA~\cite{karmanov2024efficient} forms the cache model by collecting the most reliable test features and their pseudo labels. \textit{Beyond buffering data features and their (pseudo) labels, our work trains a more succinct episodic memory by learning class-specific prompts, which encode domain-specific knowledge and enable more effective utilization of pre-trained frozen models.}

\textbf{Class-aware prompt learning}. Most recently, two works consider class-level knowledge in prompt learning as well. However, PromptSync~\cite{khandelwal2024promptsync} proposes a class-aware prototype alignment technique by aligning the prototype of test samples with the pre-computed class prototypes of the proxy source dataset. TCP~\cite{yao2024tcp} focuses on learning a class-aware textual classifier by inserting the
class-level textual tokens into the feature layer of the text encoder. \textit{Different from them, our method directly encodes prior knowledge about the class into learnable prompts from few-shot samples and builds a prompt-matching mechanism to query class-specific prompts for test data.}

\section{Method}

\subsection{Preliminaries}

In this section, we briefly introduce the VLMs, i.e., CLIP~\cite{radford2021learning}, and recap two
mainstream approaches for parameter-efficient fine-tuning on VLMs, i.e., train-time and test-time prompt tuning~\cite{zhou2022learning,zhou2022conditional,shu2022test,abdul2024align}.

\par\textbf{Contrastive language-image pre-training.}
CLIP \cite{radford2021learning} is designed to procure visual representations via natural language supervision in a contrastive learning setting. It uses 400 million image-text pairs to train the visual and text encoders, where image features from an image encoder $\mathbf{E}_{\text{visual}}(\cdot)$ and text features from a text encoder $\mathbf{E}_{\text{text}}(\cdot)$ are
aligned within a unified embedding space. For zero-shot inference, 
CLIP can classify a query image $x$ into $C$ possible
categories by matching the image feature $\textbf{f}$ with text features 
$\{\textbf{w}_c\}_{c=1}^{C}$.
The predicted probability
for the class $y_c$ is formulated as 
$p(y_c|x) = \frac{\exp(\langle  \textbf{w}_c, \textbf{f} \rangle / \tau)}{\sum_{c=1}^C \exp(\langle  \textbf{w}_c, \textbf{f}, \rangle / \tau)}$, where $\langle \cdot, \cdot \rangle$ represents the cosine similarity, and $\tau$ is the
temperature parameter. Then, its prediction probability on all classes $C$ can be denoted by $\textbf{p}(x)=\{p(y_1|x), p(y_2|x), \cdots, p(y_C|x)\}$.

\par\textbf{Train-time prompt learning.}
Prompt learning is an emerging technique to design task-specific prompts, which retains the feature representations of a pre-trained model and re-purposes them for downstream data. Different from prompt engineering~\cite{liu2023pre} that manually designs text prompts for downstream tasks, train-time prompt learning~\cite{zhou2022learning,zhou2022conditional} explores more effective learnable text or visual prompts with few labeled data. Here, we present a text prompt learning baseline that introduces learnable word vectors for class names, denoted as $P\in \mathbb{R}^{L \times D}$, where $L$ is the embedding size and $D$ is the number of tokens. The learned prompts replace hand-crafted text prompt templates by concatenating learned text prompts with 
class names as follows:
\begin{align}
\{P; \mathcal{Y}\} = \{P_1, P_2, \ldots, P_m; \mathcal{Y}\},
\end{align}
where $\mathcal{Y}$ represents class names. 
To adapt CLIP \cite{radford2021learning} to a specific image recognition task,
text prompts $P$ are optimized with the cross-entropy loss, $\mathcal{L}_{CE}$, which is defined as follows:
\begin{align}
P^{*} = \arg\min_{P} \mathbb{E}_{(x,y) \sim D} [\mathcal{L}_{CE}(\mathbf{F}_P(x), y)],
\end{align}
where $\mathbf{F}_P(x) = \langle \left(\mathbf{E}_{\text{text}}\left(\{P; \mathcal{Y}\}\right), \mathbf{E}_{\text{visual}}(x)\right \rangle$ and $D$ indicates a set with few labeled images, i.e., few-shot samples. It is noted that visual and text encoders of CLIP \cite{radford2021learning} are frozen throughout the overall training process. Despite its effectiveness, this line of methods is susceptible to overfitting on the supervised downstream task and has limited generalization towards new domains.

\begin{figure*}[t]
  \centering
\includegraphics[width=0.9\textwidth]{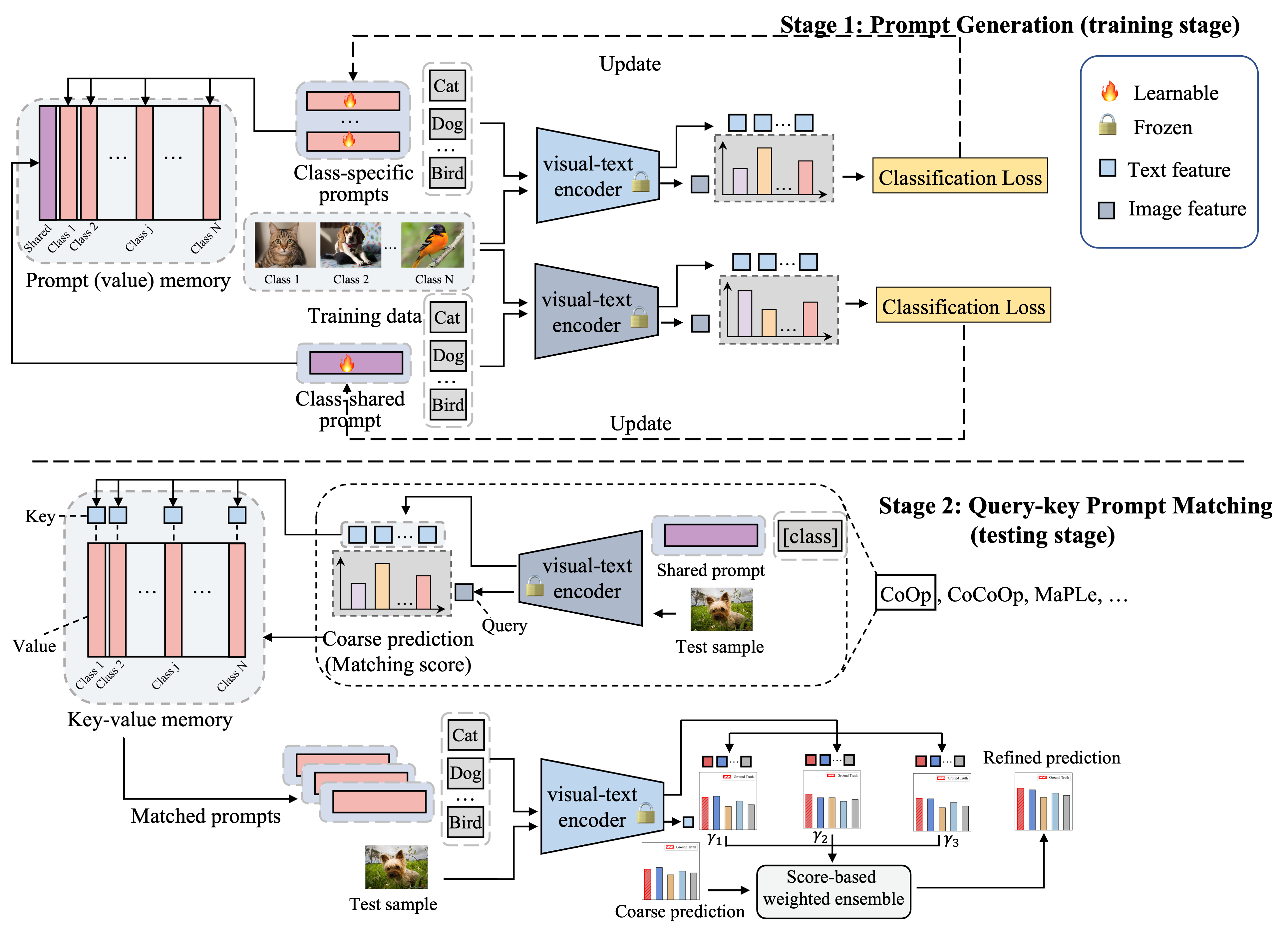}
  \caption{Workflow of the proposed CAKI. CAKI leverages class-aware prompt learning to generate
class-specific prompts and preserve these prompts as values in cache memory, which also maintains class text features as keys. With the key-value cache, our CAKI can match test samples with a set of class-specific prompts using the domain-specific information provided by existing methods (e.g., CoOp). Further, we ensemble the model predictions across different prompts based on the matching
scores, and combine these ensembled prediction with coarse prediction to produce the refined prediction for each test sample.
}\label{fig:abc}
\end{figure*}

\par\textbf{Test-time prompt learning.}  To enhance the model generalization on new data distributions, test-time prompt tuning (TPT)~\cite{shu2022test,abdul2024align} offers a solution by learning adaptive prompts on the fly with a single test sample in a zero-shot manner. This can be achieved with several optimization steps on an unsupervised objective that involves only unlabeled test samples. 
As a result, TPT flexibly provides the model with the context prompt tailored to each single test sample, thereby alleviating the distribution shift across domains. Given a single test sample $x_{\text{test}}$ without label information, the optimization objective of TPT is:
\begin{align}
P^* = \arg\min_{P} \mathcal{L}_{CE}(\mathbf{F}_P(x_{\text{test}})).\vspace{-8pt}
\end{align}
Note that, unlike Equation (2), TPT does not require any labels or additional data beyond the zero-shot test sample.

\subsection{Class-Aware Knowledge Injection}

Train-time adaptation methods use limited data to learn  class-shared prompt for test data, while test-time methods tailor instance-specific prompt to adapt pre-trained models to each individual test sample. Despite their effectiveness, these methods are unable to model the class distribution explicitly, therefore only exhibiting suboptimal performance for downstream tasks.
We argue that the challenge of exploring class-specific knowledge for test samples lies in accurately modeling prior knowledge about class for downstream tasks and retrieving relevant knowledge as context information for different test samples. Hence, we propose a class-aware prompt matching framework, CAKI, to handle these challenges. As depicted in Figure ~\ref{fig:abc}, CAKI develops class-aware prompt learning to model class distribution and generate class-specific prompts, which are stored in cache memory. Further, query-key prompt matching is proposed to facilitate the test-time adaption process with class-specific prompts. We elaborate on these modules below.

\subsubsection{Class-specific Prompt Generation}

To adapt a pre-trained model to unseen domains, the previous methods have used training or testing data to learn class-shared and instance-specific information about downstream tasks. Different from them, class-specific knowledge is more fine-grained and precise and can be more easily integrated with the well-learned knowledge of pre-trained models. To provide the model with class-specific knowledge, we develop class-aware prompt learning (CPL), which uses training data from the same class to learn class-specific prompts. 

Specifically, CPL includes an image encoder and a text encoder, denoted as $f(\cdot)$ and $g(\cdot)$, respectively. Let $\mathcal{D}_{\text{train}} = \{(x_{i}, y)\}_{i=1}^{N_{y}}$, $y \in \mathcal{Y} \}$ be the training data, where $x_i$ is the $i$-th data sample belonging to class $y$, $\mathcal{Y}$ is text class names, and $N_y$ is the size of data samples belonging to class $y$. Given text class names $\mathcal{Y} = \{y_1, y_2, \cdots, y_C\}$, we first prepend them with a hand-crafted prompt template $P$ = ``a photo of a" to form the prompted text inputs $\{(P; y) | (p, y_c) \text{ for } y_c \in \mathcal{Y}\}$.
We then feed images and prompted texts into visual and text encoders. This process yields image features, denoted as $\mathbf{f}$, and a series of text features, denoted as $\mathbf{w}_c^P$ for each class $y_c$ ranging from 1 to $C$. The probability for the $c$-th class is obtained via
\begin{align}
p(y_c^{P_c}|x) = \frac{\exp(\langle \mathbf{w}_c^{P_c}, \mathbf{f} \rangle / \tau)}{\sum_{j=1}^{C} \exp(\langle \mathbf{w}_j^{P_c}, \mathbf{f} \rangle / \tau)}.
\end{align}
Here $P_C$ contains learnable context vectors that are initialized with the pre-trained word embeddings of ``a photo of a'' as a starting point, and then learned with few-shot image-label pairs belonging to class $y_c$.
The goal of CPL is to identify an appropriate template as a class-specific prompt $P_c$ to narrow the distribution shift. 
The optimization of $P_c$ is facilitated through prompt tuning, which is formulated as 
\begin{align}
P_c = \arg\min_{P_c} \mathbb{E}_{(x,y) \sim \mathcal{D}_{\text{train}}}-\log p(y_c^P|x),
\end{align}
where $\mathcal{D}_{\text{train}}=\{x_i, y_c\}_{i=1}^{N_{y_c}}$ are few-shot image-label pairs belonging to class $y_c$.

The learned class-specific prompts are not only capable of storing the class knowledge about the target domain but also provide the model with class-specific information. To precisely and efficiently store this knowledge, we establish a key-value cache by maintaining class-specific prompts as values and the corresponding text class
features as keys:
\begin{align}
\mathcal{P} = (\left\{ (k_1, P_1), (k_2, P_2), \ldots, (k_C, P_C) \right\}),
\end{align}
where $k_c \in \mathbb{R}^{D_k}$ refers to the textual embeddings generated by encoding the original class names using the text encoder, $C$ is the number of class, $P_c \in \mathbb{R}^{L \times D}
$ is a class-specific prompt with a token length
$L_p$ and embedding size $D$. Meanwhile, we employ existing train-time and test-time methods to encode domain-specific information of downstream tasks into prompts. For instance, in this paper, we use CoOp as an example and store its class-shared prompts in cache memory, as shown in Figure~\ref{fig:abc}.

\subsubsection{Query-key Prompt Matching}

During the testing phase, we cannot directly select the appropriate class prompts because the class of the test samples are unknown. 
Images from different classes often exhibit significant similarities in certain visual patterns. For instance, cats and tigers share notable similarities in their shapes and markings. Motivated by this observation, we propose query-key prompt matching that retrieves relevant class-specific prompts to instruct the model prediction for different test instances, following a coarse-to-fine manner. 
\par
\textbf{Coarse-grained prediction.} Let $\gamma: \mathbb{R}^{D_k} \times \mathbb{R}^{D_k} \to \mathbb{R}$ be a score function to quantify the matching between the test image feature and class-based text features. We associate text features as key to the learnable prompt value:
\(\left\{ (k_1, P_1), (k_2, P_2), \ldots, (k_C, P_C) \right\}\), where \(k_i \in \mathbb{R}^{D_k}\). Ideally, we allow the test sample itself to decide which prompts should be chosen through prompt matching. Given an input image $x$, we use image feature $q(\mathbf{x})=f(x)$ to lookup the top-$K$ keys $\mathcal{K}_x$ by solving the following objective:
\begin{align}
\mathcal{K}_x, \mathcal{P}_x = \{ (k_{s_i}, P_{s_i}) | k_{s_i} \in \underset{\{s_i\}_{i=1}^K \subseteq [1,C]}{\mathrm{argmax}}\, \text{top}\, K [\gamma (q(x), k_{s_i})] \},
\end{align}
where $\mathcal{P}_x$ represents a subset of top-$K$ prompts already selected for $x$ from key-value memory $\mathcal{P}$. This query-key strategy can retrieve a set of prompts about test sample, which instruct model to make predictions with class-specific information. As a result, our approach can effectively adapt VLMs to test data with query-key prompt matching.

\par
\textbf{Fine-grained prediction.} Given a test sample \(x_t\), our proposed CAKI makes predictions based on the top-\(K\) most class-relevant prompts $\mathcal{K}_{x_t}=\{P_{s_1}, P_{s_2}, \ldots, P_{s_K}\}$, which are dynamically selected from key-value cache via prompt matching. The inference process involves aggregating $K$ model predictions $\textbf{p}^{(s_i)}(x_t)$ with the matching scores $\gamma (q(x_t), k_{s_i})$ as weights. This can be formulated as:\vspace{-6pt}
\begin{align}
\textbf{p}^{(s)}(x_t) = \sum_{i=1}^K [\gamma (q(x_t), k_{s_i}) \cdot \textbf{p}^{(s_i)}].
\end{align}\vspace{-0.25cm}

The proposed CAKI is a plug-and-play approach, which is compatible with most existing train-time and test-time prompt learning methods. This can be achieved by 
\begin{align}
\textbf{p}^{(*)}{(x_t)} = \textbf{p}(x_t) + \beta\textbf{p}^{(s)}(x_t),
\end{align}\vspace{-0.05cm}
where $\textbf{p}(x_t)$ represents the coarse model prediction obtained through either train-time or test-time prompt learning methods, and $\beta$ is a weight parameter that balances the importance between coarse and fine model predictions.

\par\textbf{Discussion.}
Our CAKI leverages different class prompts to model the knowledge distribution for different classes. During the testing stage, it utilizes existing methods to predict the class of the test sample, subsequently retrieving relevant class prompts to instruct the model's final prediction.
L2P~\cite{wang2022learning} also uses a set of prompts to learn and store task knowledge, while these prompts are still shared for some relevant tasks during the training stage. Different from L2P~\cite{wang2022learning}, our design explicitly decouples different class knowledge, thus reducing the interference between them during optimization. On the other hand, our method introduces class-specific prompts into test-time adaptation for zero-shot and few-shot vision recognition tasks, while L2P utilizes a shared prompt pool to manage task-shared and task-specific knowledge for continual learning. We also compare our CAKI with L2P~\cite{wang2022learning} by directly using a shared prompt pool to model class-share and class-specific knowledge. As shown in Table~\ref{tab:compare l2p} in the experiment results, its performance is largely inferior to our CAKI with class-specific prompts.

\section{Experiments}

\subsection{Datasets}
The proposed CAKI is evaluated on 10 downstream datasets:
\textbf{Caltech-101}~\cite{fei2004learning}: A widely-used dataset containing images across 101 object categories, often used for benchmarking object recognition algorithms. 
\textbf{Oxford-Pets}~\cite{parkhi2012cats}: A dataset consisting of images of cats and dogs, with a total of 37 different breeds. 
\textbf{Stanford Cars}~\cite{krause20133d}: A dataset containing images of cars, categorized into 196 classes based on year, make, and model. 
\textbf{Oxford-Flowers102}~\cite{nilsback2008automated}: A dataset with 102 flower categories, each containing between 40 and 258 images.
\textbf{Food-101}~\cite{bossard2014food}: A dataset featuring 101 food categories, each with 1,000 images.
\textbf{FGVC Aircraft}~\cite{maji2013fine}: A dataset focused on aircraft images, classified into 100 different categories. 
\textbf{EuroSAT}~\cite{helber2019eurosat}: A dataset consisting of satellite images of land use and land cover, categorized into 10 classes.
\textbf{SUN-397}~\cite{xiao2010sun}: A dataset containing images of scenes, categorized into 397 classes.
\textbf{Describable Textures (DTD)}~\cite{cimpoi2014describing}: A dataset containing images of various textures, with 47 classes. 
\textbf{UCF-101}~\cite{soomro2012ucf101}: A dataset of human actions in videos, categorized into 101 action categories.

\subsection{Benchmark settings}

We conduct a comprehensive evaluation of our approach and compare it with other methods across three benchmark settings.

\textbf{Base-to-novel class generalization:} 
Following MaPLe~\cite{khattak2023maple} and PromptSRC~\cite{khattak2023self}, we
split the datasets into base and novel classes.
To evaluate the generalizability of our approach within a dataset, we train the model only on base classes and then evaluate it on both base classes and novel classes. For each dataset, we test the model’s generalization for different $K$-shots per base class, where $K$ = 1, 4, 16. In this setting, we also verify whether our approach is effectively compatible with both train-time and test-time prompt learning under 1-shot labeled images, enabling us to fully leverage their advantages.

\textbf{Few-shot learning:} Following TCP~\cite{yao2024tcp}, we incorporate few-shot learning setting to evaluate the model's learning capacity under limited supervision. This allows us to verify whether our approach effectively learns both class-specific and class-agnostic knowledge. For each dataset, we assess the model's recognition ability using 16-shot labeled images.

\textbf{Domain generalization:} We evaluate the robustness of our method on out-of-distribution datasets. Following MaPLe~\cite{khattak2023maple} and PromptSRC~\cite{khattak2023self}, we train our model on ImageNet dataset and test the trained model on four other ImageNet datasets that contain various types of domain shifts.

\subsection{Implementation Details}
Our approach consists of two stages. The training stage learns class-specific and shared prompts to store knowledge specific to the target domain in cache memory, and the testing stage queries class-specific prompts for different test samples, which allows pre-trained model to perform instance-wise adaptation. Specifically, our approach learns class-specific prompts by projecting image-label pairs from each training class into text input space. We also use existing methods to learn global prompts~\cite{zhou2022learning} or instant-specific prompts~\cite{shu2022test} to store domain-specific information about target distribution. These prompts have four prompt tokens, which are initialized with the pre-trained word embeddings of the template ``a photo of a'' and then refined through 5 optimization steps on a pre-trained ViT-B/16 CLIP model. All prompts are trained for 5 epochs with a batch size of 1, and a learning rate of 0.005 via AdamW optimizer on a single NVIDIA RTX 3090 GPU with 24 GB of memory. The temperature parameter $\tau$ is set to a default value of 1. During testing inference stage, we use the pre-trained CLIP and its global prompts to extract image and text features as query and key, and then lookup top-$K$ prompts from key-value memory based on image-text similarities (i.e., coarser model predictions). When the model makes predictions on test data, these prompts are used to retrieve the class-specific knowledge from the pre-trained model and generate class-aware predictions. Finally, we combine these refined predictions with the coarser predictions to produce the final model prediction for each test sample. We report base and novel class accuracies and their harmonic mean (HM). By default, unless otherwise specified, we use CoOp to generate coarse predictions and report accuracy on the base classes for base-to-novel setting. Due to space constraints, we have rounded all experimental results to one decimal place.
For zero-shot tasks, we establish the prompt cache using few-shot data from the first half classes. The class-specific prompts in the cache are utilized to recognize test data from the remaining half classes. All the methods were conducted three times with different random seeds and
the means and the standard deviations of the overall classification
accuracies (i.e., HM) are reported.

\begin{table*}[htbp]
  \centering
  \caption{Comparison with train-time prompt learning methods for base-to-novel generalization in 1-shot setting. All experimental results are based on our re-implementation of the code released by the authors. Here we present base and novel accuracies, along with their harmonic mean (HM). The corresponding mean and standard deviation of HM also reported.}\vspace{-3pt}
  \scalebox{0.44}{
    \begin{tabular}{l|ccc|ccc|ccc|ccc|ccc}
    \toprule
    \toprule
    \multirow{2}[3]{*}{Method} & \multicolumn{3}{c|}{Flower102} & \multicolumn{3}{c|}{DTD} & \multicolumn{3}{c|}{Pets} & \multicolumn{3}{c|}{Cars} & \multicolumn{3}{c}{UCF101} \\
\cmidrule{2-16}          & Base  & Novel & HM    & Base  & Novel & HM    & Base  & Novel & HM    & Base  & Novel & HM    & Base  & Novel & HM \\
    \toprule
    
    CLIP  & 72.2  & 77.8  & 74.9  & 53.4  & 59.9  & 56.5  & 91.3  & 97.4  & 94.3  & 63.2  & 74.8  & 68.5  & 70.3  & 77.4  & 73.9 \\
    CoOp  & 81.6  & 67.8  & 74.1$\pm$0.2 & 56.3  & 56.3  & 56.3$\pm$0.1 & 93.6  & 97.2  & 95.4$\pm$0.2 & 66.1  & 73.4  & 69.6$\pm$0.2 & 76.5  & 74.4  & 75.6$\pm$0.5 \\
    CoOp+Ours & 84.2  & 71.6  & 77.4$\pm$0.4(+3.3) & 57.2  & 56.8  & 57$\pm$0.2(+0.7) & 94.4  & 98.1  & 96.2$\pm$0.1(+0.8) & 67.4  & 74.5  & 70.7$\pm$0.1(+1.1) & 77.5  & 77.3  & 77.4$\pm$0.2(+1.8) \\
    CoCoOp & 73.5  & 76.5  & 75.0$\pm$0.2 & 60.6  & 53.4  & 56.8$\pm$0.1 & 93.6  & 97.1  & 95.3$\pm$0.0 & 65.2  & 75.1  & 69.8$\pm$0.2 & 71.7  & 75.5  & 73.6$\pm$0.2 \\
    CoCoOp+Ours & 75.2  & 78.7  & 76.9$\pm$0.1(+1.9) & 61.4  & 54.4  & 57.7$\pm$0.2(+0.9) & 93.8  & 97.2  & 95.5$\pm$0.1(+0.2) & 66.3  & 75.1  & 70.4$\pm$0.0(+0.6) & 72.4  & 76.2  & 74.3$\pm$0.2(+0.7) \\
    MaPLe & 75.6  & 76.8  & 76.2$\pm$0.3 & 60.6  & 56.8  & 58.6$\pm$0.1 & 93.9  & 96.8  & 95.3$\pm$0.1 & 65.3  & 74.2  & 69.5$\pm$0.2 & 75.5  & 74.2  & 74.8$\pm$0.1 \\
    MaPLe+Ours & 77.6  & 77.3  & 77.5$\pm$0.1(+1.3) & 61.2  & 57.3  & 59.2$\pm$0.2(+0.6) & 94.2  & 97.3  & 95.7$\pm$0.2(+0.4) & 66.2  & 74.9  & 70.3$\pm$0.1(+0.8) & 76.6  & 75.8  & 76.2$\pm$0.2(+1.4) \\
    PromptSRC & 85.6  & 76.3  & 80.7$\pm$0.3 & 66.1  & 60.0    & 62.9$\pm$0.3 & 93.5  & 97.2  & 95.3$\pm$0.1 & 68.4  & 74.3  & 71.2$\pm$0.0 & 77.2  & 79.6  & 78.4$\pm$0.2 \\
    PromptSRC+Ours & 86.7  & 77.7  & 82.0$\pm$0.2+1.3) & 66.9  & 61.4  & 64.0$\pm$0.1(+1.1) & 94.1  & 97.4  & 95.7$\pm$0.0(+0.4) & 68.5  & 74.8  & 71.5$\pm$0.3(+0.3) & 78.3  & 79.9  & 79.1$\pm$0.1(+0.7) \\
    TCP   & 86.6  & 76.7  & 81.4$\pm$0.2 & 66.1  & 57.1  & 61.3$\pm$0.0 & 92.6  & 97.0    & 94.8$\pm$0.2 & 66.5  & 72.3  & 69.3$\pm$0.1 & 78.0    & 78.8  & 78.4$\pm$0.1 \\
    TCP+Ours & 88.8  & 77.8  & 83.0$\pm$0.3(+1.6) & 66.8  & 58.6  & 62.4$\pm$0.3(+1.1) & 93.6  & 97.7  & 95.6$\pm$0.3(+0.8) & 68.3  & 73.5  & 70.8$\pm$0.2(+1.5) & 79.3  & 79.4  & 79.4$\pm$0.3(+1.0) \\
    GalLoP & 86.6  & 76.3  & 81.1$\pm$0.4 & 66.4  & 59.9  & 63.0$\pm$0.1 & 93.3  & 97.5  & 95.4$\pm$0.1 & 69.5  & 74.1  & 71.7$\pm$0.2 & 77.4  & 80.3  & 78.8$\pm$0.1 \\
    GalLoP+Ours & 87.3  & 77.5  & 82.1$\pm$0.1(+1.0) & 67.5  & 61.6  & 64.4$\pm$0.1(+1.4) & 93.8  & 98.3  & 96.0$\pm$0.2(+0.6) & 70.6  & 74.4  & 72.5$\pm$0.1(+0.8) & 78.9  & 82.0    & 80.4$\pm$0.2(+1.6) \\
    \midrule
    \midrule
       \multirow{2}[3]{*}{Method}   & \multicolumn{3}{c|}{Food101} & \multicolumn{3}{c|}{SUN397} & \multicolumn{3}{c|}{Caltech101} & \multicolumn{3}{c|}{Aircraft} & \multicolumn{3}{c}{EuroSAT} \\
\cmidrule{2-16}          & Base  & Novel & HM    & Base  & Novel & HM    & Base  & Novel & HM    & Base  & Novel & HM    & Base  & Novel & HM \\
\toprule
    CLIP  & 90.1  & 91.3  & 90.7  & 69.6  & 75.4  & 72.4  & 96.6  & 94.2  & 95.4  & 27.2  & 36.4  & 31.3  & 56.4  & 64.3  & 60.1 \\
    CoOp  & 88.2  & 90.9  & 89.5$\pm$0.1 & 70.3  & 70.4  & 70.4$\pm$0.5 & 96.8  & 94.8  & 95.8$\pm$0.0 & 26.7  & 33.8  & 29.8$\pm$0.5 & 70.4  & 68.7  & 69.5$\pm$0.3 \\
    CoOp+Ours & 89.3  & 91.5  & 90.4$\pm$0.2(+0.9) & 73.2  & 72.7  & 73.0$\pm$0.3(+2.6) & 97.2  & 95.1  & 96.1$\pm$0.0(+0.3) & 30.3  & 35.9  & 32.9$\pm$0.4(+3.1) & 72.4  & 70.4  & 71.4$\pm$0.2(+1.9) \\
    CoCoOp & 89.2  & 90.7  & 89.9$\pm$0.1 & 73.2  & 76.3  & 74.7$\pm$0.2 & 96.3  & 95.6  & 96.0$\pm$0.1 & 29.4  & 35.8  & 32.3$\pm$0.2 & 65.6  & 75.4  & 70.2$\pm$0.5 \\
    CoCoOp+Ours & 89.9  & 91.6  & 90.7$\pm$0.1(+0.8) & 73.7  & 77.4  & 75.5$\pm$0.1(+0.8) & 96.3  & 96.1  & 96.2$\pm$0.2(+0.2) & 30.4  & 38.1  & 33.8$\pm$0.3(+1.5) & 70.4  & 75.3  & 72.8$\pm$0.4(+2.6) \\
    MaPLe & 87.9  & 90.4  & 89.1$\pm$0.3 & 74.4  & 77.6  & 76.0$\pm$0.2 & 96.3  & 95.6  & 96$\pm$0.1 & 30.7  & 34.1  & 32.3$\pm$0.4 & 53.8  & 60.6  & 57.0$\pm$0.3 \\
    MaPLe+Ours & 89.6	 & 91.3  & 91.3$\pm$0.4(+2.2) & 76.5  & 78.1  & 77.3$\pm$0.3(+1.3) & 96.8  & 96.1  & 96.5$\pm$0.2(+0.5) & 32.4  & 36.5  & 34.3$\pm$0.4(+2.0) & 55.2  & 62.6  & 58.8$\pm$0.1(+1.8) \\
    PromptSRC & 87.8  & 91.4  & 89.6$\pm$0.2 & 75.6  & 76.8  & 76.2$\pm$0.3 & 97.3  & 95.3  & 96.3$\pm$0.2 & 15.1  & 8.3   & 10.7$\pm$0.5 & 68.8  & 69.2  & 69$\pm$0.2 \\
    PromptSRC+Ours & 89.5  & 91.2  & 90.3$\pm$0.2(+0.7) & 76.7  & 77.4  & 77.1$\pm$0.2(+0.9) & 98.1  & 96.4  & 97.2$\pm$0.3(+0.9) & 23.7  & 23.2  & 23.5$\pm$0.2(+12.8) & 70.3  & 69.5  & 69.9$\pm$0.1(+0.9) \\
    TCP   & 90.0    & 91.2  & 90.6$\pm$0.3 & 75.0    & 77.5  & 76.2$\pm$0.2 & 97.3  & 95.1  & 96.2$\pm$0.3 & 30.4  & 34.4  & 32.3$\pm$0.4 & 77.0    & 74.7  & 75.8$\pm$0.0 \\
    TCP+Ours & 91.1  & 92.2  & 91.6$\pm$0.2(+1.0) & 76.4  & 78.8  & 77.6$\pm$0.4(+1.4) & 98.2  & 95.9  & 97$\pm$0.1(+0.8) & 33.5  & 36.1  & 34.8$\pm$0.2(+2.5) & 78.4  & 74.7  & 76.5$\pm$0.2(+0.7) \\
    GalLoP & 86.5  & 91.7  & 89.0$\pm$0.4 & 76.3  & 77.9  & 77.1$\pm$0.2 & 98.3  & 95.1  & 96.7$\pm$0.2 & 28.9  & 27.6  & 28.2$\pm$0.2 & 72.5  & 70.6  & 71.5$\pm$0.1 \\
    GalLoP+Ours & 88.8  & 91.9  & 90.3$\pm$0.3(+1.3) & 77.3  & 78.2  & 77.8$\pm$0.0(+0.7) & 98.9  & 96.4  & 97.6$\pm$0.1(+0.9) & 30.9  & 30.1  & 30.5$\pm$0.3(+2.3) & 72.5  & 71.7  & 72.1$\pm$0.2(+0.6) \\
    \bottomrule
    \bottomrule
    \end{tabular}}%
  \label{tab:few_shot}\vspace{-4pt}
\end{table*}%

\subsection{Comparison Results}

\subsubsection{Base-to-Novel Generalization}

\textbf{Comparison with train-time prompt learning.} 
As a train-time method, train-time prompt learning can provide domain-specific information for pre-trained model, which can be incorporated into our approach to derive coarse model predictions (i.e., query-key matching scores). Here we compare our CAKI with five representative train-time methods in the 1-shot setting: CoOp~\cite{zhou2022learning}, CoCoOp~\cite{zhou2022conditional}, MaPLe~\cite{khattak2023maple}, PromptSRC~\cite{khattak2023self}, TCP~\cite{yao2024tcp}, and GalLoP~\cite{lafon2024gallop}.
As shown in Table~\ref{tab:few_shot}, our proposed CAKI yields better overall performance across 10 datasets over five representative methods. 
For example, CAKI achieves a HM improvement of 1.3\%  over MaPLe on the Flower102 dataset. This is attributed to train-time prompt learning's ability to distill general knowledge from the CLIP model for downstream tasks. Our CAKI builds on this to further explore prior knowledge about class, which is used to acquire finer model predictions.

\begin{table*}[htbp]
  \centering
\caption{Comparison with train-time prompt learning methods for base-to-novel generalization in 4-shot setting. All experimental results are based on our re-implementation of the code released by the authors. Here we present base and novel accuracies, along with their harmonic mean (HM). The corresponding mean and standard deviation of HM also reported.}
\scalebox{0.44}{
    \begin{tabular}{l|ccc|ccc|ccc|ccc|ccc}
    \toprule
    \toprule
    \multirow{2}[4]{*}{Method} & \multicolumn{3}{c}{Flower102} & \multicolumn{3}{c|}{DTD} & \multicolumn{3}{c}{Pets} & \multicolumn{3}{c}{Cars} & \multicolumn{3}{c}{UCF101} \\
\cmidrule{2-16}          & Base  & Novel & HM    & Base  & Novel & HM    & Base  & Novel & HM    & Base  & Novel & HM    & Base  & Novel & HM \\
    \midrule
    CLIP  & 72.0    & 77.9  & 74.8  & 53.4  & 60.0    & 56.5  & 91.1  & 97.2  & 94.1  & 63.3  & 74.8  & 68.6  & 70.3  & 77.7  & 73.8 \\
    CoOp  & 91.3  & 68.3  & 78.1$\pm$0.2 & 66.5  & 53.4  & 59.2$\pm$0.2 & 95.2  & 97.5  & 96.3$\pm$0.1 & 70.4  & 72.4  & 71.4$\pm$0.2 & 80.5  & 75.7  & 78.0$\pm$0.2 \\
    CoOp+Ours & 91.4  & 75.6  & 82.8$\pm$0.3(+4.7) & 67.7  & 55.2  & 60.8$\pm$0.3(+1.6) & 95.4  & 97.7  & 96.5$\pm$0.2(+0.2) & 71.3  & 73.6  & 72.4$\pm$0.1(+1.0) & 81.1  & 78.3  & 79.7$\pm$0.3(+1.7) \\
    CoCoOp & 80.5  & 73.3  & 76.7$\pm$0.4 & 63.8  & 55.1  & 59.1$\pm$0.1 & 94.3  & 97.9  & 96.1$\pm$0.0 & 65.6  & 76.3  & 70.6$\pm$0.0 & 75.2  & 75.6  & 75.4$\pm$0.1 \\
    CoCoOp+Ours & 81.3  & 76.4  & 78.8$\pm$0.1(+2.1) & 64.5  & 56.4  & 60.2$\pm$0.2(+1.1) & 94.5  & 98.3  & 96.4$\pm$0.1(+0.3) & 66.2  & 75.6  & 70.6$\pm$0.1(+0.0) & 75.6  & 77.4  & 76.5$\pm$0.2(+1.1) \\
    MaPLe & 87.3  & 75.1  & 80.7$\pm$0.3 & 67.4  & 50.4  & 57.7$\pm$0.2 & 94.7  & 97.5  & 96.1$\pm$0.3 & 69.3  & 73.4  & 71.3$\pm$0.1 & 79.7  & 78.9  & 79.3$\pm$0.1 \\
    MaPLe+Ours & 89.2  & 77.2  & 82.8$\pm$0.2(+2.1) & 68.4  & 51.8  & 59$\pm$0.1(+1.3) & 94.8  & 97.7  & 96.2$\pm$0.2(+0.1) & 69.5  & 73.7  & 71.5$\pm$0.0(+0.2) & 79.8  & 79.9  & 79.9$\pm$0.2(+0.6) \\
    PromptSRC & 94.7  & 76.3  & 84.5$\pm$0.1 & 74.4  & 57.7  & 65$\pm$0.1 & 95.9  & 97.2  & 96.5$\pm$0.0 & 73.6  & 74.5  & 74.1$\pm$0.1 & 84.9  & 77.6  & 81.1$\pm$0.2 \\
    \multicolumn{1}{c|}{PromptSRC+Ours} & 94.6  & 77.7  & 85.3$\pm$0.1(+0.8) & 75.1  & 58.3  & 65.6$\pm$0.2(+0.6) & 96.0    & 97.4  & 96.7$\pm$0.1(+0.2) & 73.7  & 75.1  & 74.4$\pm$0.1(+0.3) & 85.1  & 78.5  & 81.7$\pm$0.0(+0.6) \\
    TCP   & 95.2  & 75.8  & 84.4$\pm$0.2 & 72.5  & 55.5  & 62.9$\pm$0.2 & 94.8  & 97.2  & 96.0$\pm$0.2 & 71.4  & 74.1  & 72.7$\pm$0.1 & 83.6  & 79.2  & 81.3$\pm$0.4 \\
    TCP+Ours & 96.3  & 76.4  & 85.2$\pm$0.2(+0.8) & 73.6  & 57.6  & 64.6$\pm$0.3(+1.7) & 95.2  & 97.9  & 96.5$\pm$0.1(+0.5) & 72.5  & 74.7  & 73.6$\pm$0.2(+0.9) & 84.9  & 80.7  & 82.8$\pm$0.3(+1.5) \\
    GalLoP & 95.3  & 76.0    & 84.6$\pm$0.3 & 75.6  & 56.7  & 64.8$\pm$0.1 & 95.8  & 97.4  & 96.6$\pm$0.0 & 74.1  & 75.2  & 74.7$\pm$0.0 & 85.6  & 78.8  & 82.1$\pm$0.3 \\
    GalLoP+Ours & 96.5  & 76.4  & 85.3$\pm$0.2(+0.7) & 76.7  & 57.6  & 65.8$\pm$0.3(+1.0) & 96.2  & 97.5  & 96.9$\pm$0.1(+0.3) & 74.2  & 75.8  & 75.0$\pm$0.1(+0.3) & 86.9  & 79.3  & 83.0$\pm$0.1(+0.9) \\
    \midrule
    \midrule
    \multirow{2}[4]{*}{Method} & \multicolumn{3}{c|}{Food101} & \multicolumn{3}{c|}{SUN397} & \multicolumn{3}{c}{Caltech101} & \multicolumn{3}{c|}{Aircraft} & \multicolumn{3}{c}{EuroSAT} \\
\cmidrule{2-16}          & Base  & Novel &       & Base  & Novel & HM    & Base  & Novel & HM    & Base  & Novel & HM    & Base  & Novel & HM \\
    \midrule
    CLIP  & 90.2  & 91.4  & 90.8  & 69.6  & 75.5  & 72.4  & 96.7  & 94.1  & 95.4  & 27.7  & 36.5  & 31.5  & 56.8  & 64.2  & 60.3 \\
    CoOp  & 87.3  & 82.4  & 84.8$\pm$0.3 & 70.5  & 70.4  & 70.5$\pm$0.4 & 97.9  & 94.2  & 96$\pm$0.0 & 30.6  & 28.7  & 29.6$\pm$0.5 & 80.6  & 80.1  & 80.3$\pm$0.1 \\
    CoOp+Ours & 88.3  & 83.9  & 86$\pm$0.1(+1.2) & 73.2  & 72.6  & 72.9$\pm$0.2(+2.4) & 98.1  & 94.4  & 96.2$\pm$0.1(+0.2) & 31.6  & 34.4  & 32.9$\pm$0.4(+3.3) & 81.3  & 80.5  & 80.9$\pm$0.2(+0.6) \\
    CoCoOp & 89.8  & 91.2  & 90.5$\pm$0.2 & 75.7  & 78.5  & 77.1$\pm$0.0 & 97.3  & 95.3  & 96.3$\pm$0.1 & 30.3  & 36.7  & 33.2$\pm$0.2 & 74.6  & 73.2  & 73.9$\pm$0.3 \\
    CoCoOp+Ours & 90.4  & 91.8  & 91.1$\pm$0.1(+0.6) & 76.7  & 78.4  & 77.5$\pm$0.1(+0.4) & 97.4  & 95.6  & 96.5$\pm$0.0(+0.2) & 31.7  & 37.6  & 34.4$\pm$0.3(+1.2) & 75.9  & 73.5  & 74.7$\pm$0.1(+0.8) \\
    MaPLe & 90.1  & 91.5  & 90.8$\pm$0.1 & 78.5  & 78.6  & 78.6$\pm$0.1 & 98.3  & 94.9  & 96.6$\pm$0.1 & 32.5  & 37.2  & 34.7$\pm$0.3 & 80.8  & 76.4  & 78.5$\pm$0.1 \\
    MaPLe+Ours & 90.6  & 92.0    & 91.3$\pm$0.1(+0.5) & 79.2  & 78.3  & 78.8$\pm$$\pm$0.0(+0.2) & 98.2  & 95.3  & 96.7$\pm$0.2(+0.1) & 33.6  & 38.5  & 35.9$\pm$0.2(+1.2) & 81.3  & 77.2  & 79.2$\pm$0.0(+0.7) \\
    PromptSRC & 90.3  & 91.4  & 90.8$\pm$0.2 & 80.4  & 77.7  & 79.0$\pm$0.1 & 97.8  & 94.7  & 96.2$\pm$0.0 & 34.3  & 37.1  & 35.7$\pm$0.4 & 83.7  & 78.2  & 80.9$\pm$0.3 \\
    PromptSRC+Ours & 90.4  & 91.9  & 91.1$\pm$0.2(+0.3) & 80.5  & 78.4  & 79.4$\pm$0.2(+0.4) & 98.3  & 95.4  & 96.8$\pm$0.1(+0.6) & 35    & 38.5  & 36.7$\pm$0.2(+1.0) & 85.5  & 78.6  & 81.9$\pm$0.1(+1.0) \\
    TCP   & 90.1  & 91.5  & 90.8$\pm$0.0 & 73.1  & 55.5  & 63.1$\pm$0.2 & 98.1  & 95.0    & 96.5$\pm$0.0 & 35.6  & 33.5  & 34.5$\pm$0.1 & 85.1  & 78.6  & 81.7$\pm$0.2 \\
    TCP+Ours & 90.6  & 92.2  & 91.4$\pm$0.1(+0.6) & 73.7  & 57.2  & 64.4$\pm$0.4(+1.3) & 98.5  & 95.1  & 96.7$\pm$0.1(+0.2) & 36.2  & 34.9  & 35.5$\pm$0.2(+1.0) & 86.3  & 78.8  & 82.4$\pm$0.2(+0.7) \\
    GalLoP & 90.5  & 92.5  & 91.5$\pm$0.3 & 75.6  & 56.5  & 64.7$\pm$0.1 & 97.7  & 94.6  & 96.1$\pm$0.2 & 35.9  & 37.8  & 36.8$\pm$0.3 & 84.2  & 79.2  & 81.6$\pm$0.1 \\
    GalLoP+Ours & 91.8  & 93.4  & 92.6$\pm$0.2(+1.1) & 76.4  & 57.5  & 65.6$\pm$0.2(+0.9) & 98.6  & 95.4  & 97.0$\pm$0.1(+0.9) & 36.7  & 40.0    & 38.3$\pm$0.4(+1.5) & 86.5  & 79.9  & 83.1$\pm$0.3(+1.5) \\
    \bottomrule
    \bottomrule
    \end{tabular}}%
  \label{tab:4-shot}
\end{table*}%

\begin{table*}[htbp]
  \centering
  \caption{Comparison with train-time prompt learning methods for base-to-novel generalization in 16-shot setting. Results marked with ``$*$" indicate those obtained by re-implementing these methods. Here we present base and novel accuracies, along with their harmonic mean (HM). The corresponding mean and standard deviation of HM also reported.}
    \scalebox{0.44}{
    \begin{tabular}{l|ccc|ccc|ccc|ccc|ccc}
    \toprule
    \toprule
    \multirow{2}[4]{*}{Method} & \multicolumn{3}{c|}{Flower102} & \multicolumn{3}{c|}{DTD} & \multicolumn{3}{c|}{Pets} & \multicolumn{3}{c|}{Cars} & \multicolumn{3}{c}{UCF101} \\
\cmidrule{2-16}          & Base  & Novel & HM    & Base  & Novel & HM    & Base  & Novel & HM    & Base  & Novel & HM    & Base  & Novel & HM \\
    \midrule
    CoOp  & 97.1  & 67.3  & 97.1$\pm$0.3 & 80.2  & 48.4  & 60.4$\pm$0.1 & 94.4  & 96.1  & 95.2$\pm$0.1 & 75.4  & 67.4  & 71.2$\pm$0.2 & 84.2  & 67.7  & 75.1$\pm$0.1 \\
    CoCoOp & 94.8  & 72.2  & 94.8$\pm$0.2 & 77.5  & 55.7  & 64.8$\pm$0.3 & 95.3  & 97.5  & 96.4$\pm$0.1 & 70.3  & 73.4  & 71.8$\pm$0.1 & 82.4  & 73.1  & 77.5$\pm$0.2 \\
    \midrule
    MaPLe & 95.2  & 73.7  & 95.2$\pm$0.1 & 80.2  & 57.4  & 66.9$\pm$0.2 & 95.2  & 97.3  & 96.2$\pm$0.2 & 73.5  & 74.4  & 73.9$\pm$0.1 & 83.9  & 73.6  & 78.4$\pm$0.3 \\
    MaPLe+Ours & 95.6  & 74.2  & 95.6$\pm$0.2(+0.4) & 80.3  & 58.6  & 67.8$\pm$0.2(+0.9) & 95.7  & 97.6  & 96.6$\pm$0.1(+0.4) & 74.3  & 75.5  & 74.8$\pm$0.3(+0.9) & 83.8  & 76.4  & 79.9$\pm$0.2(+1.5) \\
    PromptSRC & 97.4  & 77.3  & 97.4$\pm$0.2 & 84.5  & 61.9  & 71.5$\pm$0.1 & 95.3  & 97.6  & 96.4$\pm$0.0 & 78.5  & 76.0    & 77.2$\pm$0.1 & 86.9  & 78.5  & 82.4$\pm$0.1 \\
    \multicolumn{1}{c|}{PromptSRC+Ours} & 98.0    & 78.4  & 98$\pm$0.1(+0.6) & 85.2  & 62.3  & 72.0$\pm$0.1(+0.5) & 95.8  & 97.9  & 96.8$\pm$0.2(+0.4) & 78.3  & 75.8  & 77.0$\pm$0.2(-0.2) & 86.6  & 78.9  & 82.6$\pm$0.0(+0.2) \\
    TCP   & 98.2  & 76.5  & 98.2$\pm$0.1 & 82.3  & 58.8  & 68.6$\pm$0.2 & 95.1  & 97.7  & 96.3$\pm$0.1 & 79.4  & 74.7  & 77.0$\pm$0.1 & 86.7  & 80.2  & 83.3$\pm$0.4 \\
    TCP+Ours & 98.4  & 76.9  & 98.4$\pm$0.1(+0.2) & 82.4  & 59.9  & 69.4$\pm$0.2(+0.8) & 95.6  & 98.0    & 96.8$\pm$0.2(+0.5) & 80.6  & 75.6  & 78.0$\pm$0.2(+1.0) & 87.5  & 80.7  & 84.0$\pm$0.2(+0.7) \\
    GalLoP & 98.6  & 76.7  & 98.6$\pm$0.3 & 84.6  & 57.8  & 68.7$\pm$0.1 & 95.8  & 97.9  & 96.8$\pm$0.3 & 79.3  & 75.4  & 77.3$\pm$0.3 & 86.6  & 81.2  & 83.8$\pm$0.2 \\
    GalLoP+Ours & 99.5  & 77.6  & 99.5$\pm$0.2(+0.9) & 85.3  & 58.4  & 69.3$\pm$0.2(+0.6) & 96.8  & 98.6  & 97.7$\pm$0.4(+0.9) & 81.3  & 75.9  & 78.5$\pm$0.2(+1.2) & 88.1  & 81.7  & 84.8$\pm$0.2(+1.0) \\
    \midrule
    \midrule
    \multirow{2}[4]{*}{Method} & \multicolumn{3}{c|}{Food101} & \multicolumn{3}{c|}{SUN397} & \multicolumn{3}{c|}{Caltech101} & \multicolumn{3}{c|}{Aircraft} & \multicolumn{3}{c}{EuroSAT} \\
\cmidrule{2-16}          & Base  & Novel & HM    & Base  & Novel & HM    & Base  & Novel & HM    & Base  & Novel & HM    & Base  & Novel & HM \\
    \midrule
    CoOp  & 89.6  & 88.9  & 89.3$\pm$0.2 & 81.1  & 68.4  & 74.2$\pm$0.4 & 97.9  & 93.2  & 95.5$\pm$0.0 & 39.6  & 31.4  & 35.0$\pm$0.5 & 90.2  & 53.1  & 66.8$\pm$0.2 \\
    CoCoOp & 90.4  & 91.5  & 90.9$\pm$0.2 & 79.6  & 77.1  & 78.3$\pm$0.2 & 98.1  & 93.7  & 95.9$\pm$0.1 & 33.3  & 23.8  & 27.8$\pm$0.4 & 87.6  & 60.2  & 71.4$\pm$0.3 \\
    \midrule
    MaPLe & 90.5  & 91.8  & 91.1$\pm$0.1 & 80.5  & 78.6  & 79.5$\pm$0.1 & 98.6  & 94.3  & 96.4$\pm$0.2 & 38.5  & 35.1  & 36.7$\pm$0.2 & 93.8  & 72.4  & 81.7$\pm$0.1 \\
    MaPLe+Ours & 90.7  & 92.2  & 91.4$\pm$0.1(+0.3) & 80.6  & 78.8  & 79.7$\pm$0.2(+0.2) & 99.3  & 94.3  & 96.7$\pm$0.1(+0.3) & 39.6  & 35.5  & 37.4$\pm$0.3(+0.7) & 94.3  & 72.6  & 82.0$\pm$0.0(+0.3) \\
    PromptSRC & 90.4  & 91.5  & 91.0$\pm$0.2 & 82.4  & 78.3  & 80.3$\pm$0.2 & 98.3  & 94.1  & 96.2$\pm$0.1 & 42.6  & 37.5  & 39.9$\pm$0.2 & 91.7  & 73.2  & 81.4$\pm$0.4 \\
    PromptSRC+Ours & 90.9  & 91.8  & 91.4$\pm$0.1(+0.4) & 83.1  & 78.4  & 80.7$\pm$0.1(+0.4) & 98.4  & 94.3  & 96.3$\pm$0.0(+0.1) & 42.7  & 38.6  & 40.6$\pm$0.3(+0.7) & 92.2  & 75.6  & 83.1$\pm$0.2(+1.7) \\
    TCP   & 90.6  & 91.2  & 90.9$\pm$0.2 & 82.8  & 77.9  & 80.3$\pm$0.3 & 98.5  & 94.6  & 96.5$\pm$0.0 & 42.6  & 34.5  & 38.1$\pm$0.0 & 91.9  & 75.0    & 82.6$\pm$0.1 \\
    TCP+Ours & 90.8  & 92.5  & 91.6$\pm$0.3(+0.7) & 83.7  & 78.2  & 80.9$\pm$0.1(+0.6) & 98.6  & 95.0    & 96.8$\pm$0.1(+0.3) & 42.2  & 34.9  & 38.2$\pm$0.1(+0.1) & 92.1  & 75.4  & 82.9$\pm$0.1(+0.3) \\
    GalLoP & 90.9  & 91.5  & 91.2$\pm$0.3 & 82.5  & 78.4  & 80.4$\pm$0.2 & 97.8  & 95.2  & 96.5$\pm$0.1 & 43.6  & 35.8  & 39.3$\pm$0.3 & 92.2  & 75.6  & 83.1$\pm$0.3 \\
    GalLoP+Ours & 91.2  & 93.3  & 92.2$\pm$0.3(+1.0) & 83.4  & 79.5  & 81.4$\pm$0.1(+1.0) & 98.7  & 96.4  & 97.5$\pm$0.3(+1.0) & 44.7  & 37.5  & 40.8$\pm$0.4(+1.5) & 93.1  & 76.9  & 84.2$\pm$0.1(+1.1) \\
    \bottomrule
    \bottomrule
    \end{tabular}}%
  \label{tab:16-shot}%
\end{table*}%

We also provide the results of 4-shot and 16-shot settings in the supplementary material, which shows consistent improvements. In this experiment, we compare our CAKI method with five representative few-shot prompt learning approaches (i.e., CoOp~\cite{zhou2022learning}, CoCoOp~\cite{zhou2022conditional}, MaPLe~\cite{khattak2023maple}, PromptSRC~\cite{khattak2023self}, and TCP~\cite{yao2024tcp}) under 4-shot and 16-shot settings. As shown in Table~\ref{tab:4-shot} and Table~\ref{tab:16-shot}, CAKI outperforms existing prompt learning approaches, consistent with the findings presented in Table 1. For example, in the 4-shot setting, CAKI improves the HM accuracy of MaPLe~\cite{khattak2023maple} and TCP~\cite{yao2024tcp} by 1.3\% and 1.7\%, respectively, on the DTD dataset. 

\textbf{Summary.} These experimental results show that the mean performance of the
CAKI is better than that of existing baselines, with small standard deviations. These demonstrate that, although the absolute gains are not obvious in some benchmarks, the improvements offered by CAKI are statistically reliable.

\begin{table*}[t]
  \centering
  \caption{Evaluation of the compatibility of our CAKI framework with existing train-time and test-time prompt learning in 1-shot setting. CoOp (CoCoOp)+TPT employes the shared prompt learned by CoOp (CoCoOp) to initialize the text prompt of TPT on each test sample, while CoOp (CoCoOp)+TPT\_Pre indicates directly ensembling the predictions from both models (i.e., CoOp (CoCoOp) and TPT). All experimental results are based on our re-implementation of the code released by the authors.}\vspace{-3pt}
  \scalebox{0.49}{
    \begin{tabular}{l|ccc|ccc|ccc|ccc|ccc}
    \toprule
    \toprule
    \multirow{2}[4]{*}{Method} & \multicolumn{3}{c|}{Flower102} & \multicolumn{3}{c|}{DTD} & \multicolumn{3}{c|}{Pets} & \multicolumn{3}{c|}{Cars} & \multicolumn{3}{c}{UCF101} \\
\cmidrule{2-16}          & Base  & Novel & HM    & Base  & Novel & HM    & Base  & Novel & HM    & Base  & Novel & HM    & Base  & Novel & HM \\
    \midrule
    TPT   & 68.7  & 77.5  & 72.8  & 52.9  & 60.3  & 56.4  & 89.2  & 95.1  & 92.1  & 64.7  & 75.5  & 69.7  & 70.0  & 75.8  & 72.8  \\
    CoOp  & 82.0  & 67.4  & 74.0  & 56.8  & 56.4  & 56.6  & 93.6  & 97.3  & 95.4  & 66.0  & 73.1  & 69.4  & 76.6  & 74.6  & 75.6  \\
    CoOp+TPT & 78.1  & 65.7  & 71.4  & 56.4  & 55.6  & 56.0  & 93.7  & 94.9  & 94.3  & 66.1  & 73.4  & 69.6  & 75.4  & 73.7  & 74.5  \\
    CoOp+TPT+Ours & 80.0  & 70.8  & 75.1\small{(+3.7)}  & 57.4  & 60.4  & 58.9\small{(+2.9)}  & 94.1  & 95.4  & 94.7\small{(+0.4)}  & 66.4  & 74.0  & 70.0\small{(+0.4)}  & 76.0  & 76.6  & 76.3(\small{(+1.8)})  \\
    CoOp+TPT\_Pre & 74.2  & 77.2  & 75.7  & 55.4  & 57.4  & 56.4  & 90.2  & 96.7  & 93.3  & 67.5  & 76.1  & 71.5  & 74.3  & 79.9  & 77.0  \\
    CoOp+TPT\_Pre+Ours & 75.2  & 77.8  & 76.5\small{(+0.8)}  & 56.2  & 58.2  & 57.2\small{(+0.8)}  & 90.4  & 96.8  & 93.5\small{(+0.2)}  & 67.5  & 76.3  & 71.6\small{(+0.1)}  & 74.3  & 80.4  & 77.2\small{(+0.2)}  \\
    CoCoOp & 73.5  & 76.5  & 75.0  & 60.4  & 53.5  & 56.7  & 93.7  & 97.0  & 95.3  & 65.1  & 75.0  & 69.7  & 71.9  & 75.6  & 73.7  \\
    CoCoOp+TPT & 37.9  & 52.8  & 44.1  & 39.9  & 35.5  & 37.6  & 70.3  & 83.3  & 76.2  & 58.6  & 72.0  & 64.6  & 68.7  & 70.4  & 69.5  \\
    CoCoOp+TPT+Ours & 44.0  & 59.1  & 50.4\small{(+6.3)}  & 43.8  & 40.9  & 42.3\small{(+4.7)}  & 74.0  & 85.9  & 79.5\small{(+3.3)}  & 59.8  & 72.8  & 65.7\small{(+1.1)}  & 70.5  & 75.5  & 72.9\small{(+3.4)}  \\
    CoOpOp+TPT\_Pre & 70.8  & 79.0  & 74.7  & 54.7  & 62.3  & 58.3  & 90.0  & 96.9  & 93.3  & 66.5  & 75.7  & 70.8  & 71.6  & 76.8  & 74.1  \\
    CoOpOp+TPT\_Pre+Ours & 71.8  & 79.2  & 75.3\small{(+0.6)}  & 55.9  & 63.5  & 59.5\small{(+1.2)}  & 90.2  & 97.0  & 93.5\small{(+0.62)}    & 66.4  & 75.8  & 70.8\small{(+0.0)}    & 71.9  & 77.3  & 74.5\small{(+0.4)}    \\
    \midrule
    \midrule
    \multirow{2}[4]{*}{Method} & \multicolumn{3}{c|}{Food101} & \multicolumn{3}{c|}{SUN397} & \multicolumn{3}{c|}{Caltech101} & \multicolumn{3}{c|}{Aircraft} & \multicolumn{3}{c}{EuroSAT} \\
\cmidrule{2-16}          & Base  & Novel & HM    & Base  & Novel & HM    & Base  & Novel & HM    & Base  & Novel & HM    & Base  & Novel & HM \\
    \midrule
    TPT   & 86.6  & 91.0  & 88.7  & 71.4  & 75.9  & 73.6  & 97.4  & 86.8  & 91.8  & 25.2  & 33.6  & 28.8  & 50.1  & 65.1  & 56.6  \\
    CoOp  & 88.0  & 90.8  & 89.4  & 70.5  & 70.3  & 70.4  & 96.9  & 94.8  & 95.8  & 26.8  & 34.0  & 30.0  & 70.5  & 69.0  & 69.7  \\
    CoOp+TPT & 83.8  & 87.3  & 85.5  & 70.1  & 69.1  & 69.6  & 98.0  & 94.4  & 96.2  & 28.3  & 33.5  & 30.7  & 73.2  & 64.5  & 68.6  \\
    CoOp+TPT+Ours & 84.6  & 88.0  & 86.3\small{(+0.8)}  & 71.2  & 69.6  & 70.4\small{(+0.8)}  & 98.3  & 94.5  & 96.4\small{(+0.2)}  & 29.2  & 35.2  & 31.9\small{(+1.2)}  & 74.1  & 65.2  & 69.4\small{(+0.8)}  \\
    CoOp+TPT\_Pre & 87.9  & 91.8  & 89.8  & 74.1  & 75.9  & 75.0  & 94.0  & 92.9  & 93.4  & 27.8  & 35.2  & 31.1  & 59.1  & 67.4  & 63.0  \\
    CoOp+TPT\_Pre+Ours & 87.9  & 92.0  & 89.9\small{(+0.1)}  & 75.3  & 76.3  & 75.8\small{(+0.8)}  & 95.0  & 94.2  & 94.6\small{(+1.2)}  & 27.7  & 36.2  & 31.4\small{(+0.3)}  & 59.9  & 68.4  & 63.9\small{(+0.9)}  \\
    CoCoOp & 89.1  & 90.5  & 89.8  & 73.2  & 76.4  & 74.8  & 96.4  & 95.9  & 96.1  & 29.3  & 35.9  & 32.3  & 65.3  & 75.4  & 70.0  \\
    CoCoOp+TPT & 83.8  & 87.3  & 85.5  & 58.6  & 62.4  & 60.4  & 97.6  & 94.4  & 96.0  & 22.2  & 22.6  & 22.4  & 57.6  & 64.6  & 60.9  \\
    CoCoOp+TPT+Ours & 84.6  & 88.0  & 86.3\small{(+0.8)}  & 62.1  & 65.9  & 63.9\small{(+3.5)}  & 97.8  & 94.4  & 96.1\small{(+0.1)}  & 22.4  & 25.3  & 23.8\small{(+1.4)}  & 59.1  & 65.4  & 62.1\small{(+1.2)}  \\
    CoOpOp+TPT\_Pre & 87.9  & 91.8  & 89.8  & 73.2  & 77.5  & 75.3  & 93.0  & 92.0  & 92.5  & 28.3  & 35.5  & 31.5  & 59.5  & 70.7  & 64.6  \\
    CoOpOp+TPT\_Pre+Ours & 87.9  & 91.9  & 89.9\small{(+0.1)}  & 74.1  & 77.7  & 75.9\small{(+0.6)}  & 94.6  & 94.1  & 94.3\small{(+1.8)}  & 27.9  & 36.1  & 31.5\small{(+0.0)}  & 60.4  & 71.3  & 65.4\small{(+0.8)}  \\
    \bottomrule
    \bottomrule
    \end{tabular}}
  \label{tab:tta}
\end{table*}

\textbf{Compatibility of CAKI with existing prompt learning.} We explore the compatibility of our CAKI with both train-time and test-time prompt learning methods. \textbf{a) As for the test-time prompt learning method}, TPT~\cite{shu2022test} can learn adaptive text prompts (i.e., instance-aware prompts) for each test sample, enabling the adaptation of VLMs to new data distribution. Theoretically, test-time prompt learning can be integrated into our CAKI framework to further leverage instance-specific information. As shown in Table~\ref{tab:tta}, CAKI consistently improves the performance of TPT across all benchmark datasets. This demonstrates that CAKI is compatible with TPT, and that class-specific and instance-specific knowledge are complementary, jointly enhancing model performance. \textbf{b) As for train-time prompt learning methods} (e.g., CoOp and CoCoOp), which learn class-shared knowledge from a limited training dataset, we integrate these methods into TPT in two ways: 1) using the shared prompt to initialize the text prompt for each test sample; 2) directly ensembling the predictions from both models. Experimental results indicate that the former approach exhibits inferior performance compared to the latter in most benchmark datasets, primarily due to the overfitting of the initialized prompt to the training data, making it less suitable than the general “a photo of a” initialization. Directly ensembling the predictions from different models is a more straightforward way to leverage both class-shared and instance-specific knowledge. Regardless of which integration method is used, our approach achieves consistent improvement. For example, on the DTD dataset, our approach achieves improvements of 2.9\% and 4.7\% for CoOp+TPT and CoCoOp+TPT, respectively.

\begin{table*}[htbp]
  \centering
  \caption{Comparison of few-shot learning with 4-shot samples. The experimental results are taken from the original report in GalLoP~\cite{lafon2024gallop}. Results marked with ``$*$" indicate those obtained by re-implementing these methods.}\vspace{-3pt}
  \scalebox{0.49}{
    \begin{tabular}{lccccccccccc}
    \toprule
    \toprule
          & ImageNet  & Caltech101 & Pets  & Cars  & Flower102 & Food101 & Aircraft  & SUN397  & DTD   & EuroSAT & UCF101 \\
    \midrule
    CoOp  & 71.7  & 95.6  & 91.9  & 83.1  & 97.1  & 84.2  & 43.4  & 74.7  & 69.9  & 84.9  & 82.2 \\
    CoCoOp & 71.0    & 95.2  & 93.3  & 71.6  & 87.8  & 87.2  & 31.2  & 72.2  & 63.0    & 73.3  & 78.1 \\
    MaPLe & 72.3  & 96.0    & 92.8  & 83.6  & 97.0    & 85.3  & 48.4  & 75.5  & 71.3  & 92.3  & 85.0 \\
    PLOT  & 72.6  & 96.0    & 93.6  & 84.6  & 97.6  & 87.1  & 46.7  & 76.0    & 71.4  & 92.0    & 85.3 \\
    LoCoOp & 71.5  & 94.9  & 92.4  & 79.8  & 96.3  & 84.7  & 40.7  & 74.2  & 69.5  & 86.1  & 81.6 \\
    ProDA & 71.9  & 95.5  & 93.5  & 79.8  & 96.8  & 86.8  & 40.2  & 75.7  & 70.9  & 85.1  & 83.3 \\
    \midrule
    PromptSRC & 73.2  & 96.1  & 93.7  & 85.8  & 97.6  & 86.5  & 50.8  & 77.2  & 72.7  & 92.4  & 86.5 \\
    PromptSRC+Ours & 74.1(+0.9) & 96.6(+0.5) & 94.7(+1.0) & 86.4(+0.6) & 97.9(+0.3) & 87.8(+1.3) & 52.0(+1.2) & 77.6(+0.4) & 73.5(+0.8) & 93.1(+0.7) & 87.2(+0.7) \\
    \midrule
    TCP$^{*}$   & 72.4  & 96.2  & 91.8  & 86.3  & 96.9  & 85.4  & 49.2  & 76.4  & 71.6  & 92.2  & 86.3 \\
    TCP+Ours & 73.0(+0.6) & 96.6(+0.4) & 92.4(+0.6) & 87.1(+0.8) & 97.3(+0.4) & 86.6(+1.2) & 50.6(+1.4) & 77.2(+0.8) & 72.1(+0.5) & 92.5(+0.3) & 87.1(+0.8) \\
    \midrule
    GalLoP & 75.1  & 96.7  & 94.1  & 89.2  & 98.8  & 86.5  & 58.3  & 77.2  & 75.5  & 90.1  & 86.9 \\
    GalLoP+Ours & 75.8(+0.7) & 97.3(+0.6) & 95.2(+1.1) & 90.6(+1.4) & 99.0(+0.2) & 87.1(+0.6) & 59.1(+0.8) & 78.4(+1.2) & 76.1(+0.6) & 90.6(+0.5) & 87.3(+0.4) \\
    \midrule
    TIP-Adapter-F & 70.1  & 93.5  & 90.2  & 66.6  & 83.8  & 84.6  & 29.4  & 67.5  & 51.5  & 67.8  & 73.4 \\
    TIP-Adapter-F+Ours & 71.2(+1.1) & 94.6(+1.1) & 90.7(+0.5) & 67.8(+1.2) & 85.1(+1.3) & 85.6(+1.0) & 30.6(+1.2) & 68.9(+1.4) & 52.6(+1.1) & 68.5(+0.7) & 74.7(+1.3) \\
    \midrule
    CLIP-LoRA & 71.3  & 95.2  & 90.7  & 77.8  & 93.7  & 82.5  & 38.1  & 72.5  & 64.1  & 84.6  & 80.9 \\
   CLIP-LoRA+Ours & 72.6(+1.3) & 96.3(+1.1) & 91.6(+0.9) & 79.2(+1.4) & 95.1(+1.4) & 83.1(+0.6) & 39.6(+1.5) & 73.5(+1.0) & 65.5(+1.4) & 85.2(+0.6) & 82.1(+1.2) \\
    \bottomrule
    \bottomrule
    \end{tabular}}
  \label{tab:addlabel}%
\end{table*}%

\begin{table*}[htbp]
  \centering
  \caption{Comparison with existing methods on domain generalization. The experimental results are taken from the original report in PromptSRC~\cite{khattak2023self}. Results marked with ``$*$" indicate those obtained by re-implementing these methods.}\vspace{-3pt}
  \scalebox{0.5}{
    \begin{tabular}{lcccccc}
    \toprule
    \toprule
    \multirow{2}[4]{*}{Method} & Source &       & \multicolumn{4}{c}{Target} \\
\cmidrule{2-2}\cmidrule{4-7}          & ImageNet &       & V     & S     & A     & R \\
    \midrule
    CoOp  & 71.5 &       & 64.2  & 48.0 & 49.7 & 75.2 \\
    CoCoOp & 71.0 &       & 64.1 & 48.8 & 50.6 & 76.2 \\
    \midrule
    MaPLe$^{*}$ & 70.1 &       & 64.5 & 47.5 & 50.6 & 77.2 \\
    MaPLe+Ours & 70.7(+0.6) &       & 65.0(+0.5) & 47.7(+0.2) & 50.8(+0.2) & 77.6(+0.4) \\
    \midrule
    PromptSRC$^{*}$ & 70.9 &       & 64.9 & 48.3 & 50.3 & 78.1 \\
    PromptSRC+Ours & 71.2(+0.3) &       & 65.0(+0.1) & 48.6(+0.3) & 50.3(+0.0) & 78.5(+0.4) \\
    \bottomrule
    \bottomrule
    \end{tabular}}
  \label{tab:domain generalization}
\end{table*}

\subsubsection{Few-shot Experiments}
Following GalLoP~\cite{lafon2024gallop}, we conduct the few-shot experiments using 4-shot labeled training data and evaluate the trained model on testing data with the same class space as the training classes. The comparison between the proposed CAKI and existing methods is summarized in Table~\ref{tab:addlabel}. As shown in Table~\ref{tab:addlabel}, we can observe that the proposed method can enhance the performance of MaPLe~\cite{khattak2023maple} and PromptSRC~\cite{khattak2023self} on most benchmark datasets. For example, our approach achieves the improvements of 0.26\%, 0.35\%, and 0.59\% on ImageNet, Flower102, and DTD datasets. This demonstrate that the effectiveness of CAKI in inferring the class-aware knowledge.

\subsubsection{Domain Generalization Experiments}

In Table~\ref{tab:domain generalization}, we evaluate the direct transferability of ImageNet-trained
model on various out-of-domain datasets, and present the comparison results of our approach with existing methods. As shown in Table~\ref{tab:domain generalization}, when transferring the learned class-specific prompts to different domains, our method can achieve obvious improvement on state-of-the-art methods (i.e., MaPLe~\cite{khattak2023maple} and PromptSRC~\cite{khattak2023self}). This indicates that our CAKI framework can better enhance the generalization and robustness of existing pre-trained VLMs on datasets with domain shifts.

\subsubsection{Test-time Domain Adaptation}

Following L2C~\cite{chi2025learning}, we further evaluate CAKI on the DomainNet dataset against domain knowledge injection methods. Table~\ref{tab:test-time domain adaptation} reports the accuracy on individual domains as well as their overall averages. As shown in the table, our approach significantly surpasses VDPG and L2C, achieving average accuracy improvements of +1.1\% and +1.2\%, respectively. These results demonstrate that CAKI effectively organizes knowledge at the class level and benefit from retrieving semantically relevant information during inference.

\begin{table*}[t]
  \centering
    \caption{Comparison with existing methods on DomainNet dataset. The experimental results are taken from the original report in L2C~\cite{chi2025learning}.}
    \scalebox{0.7}{
    \begin{tabular}{lccccccc}
    \toprule
    \toprule
    Method & Clip  & Info  & Paint & Quick & Real  & Sketch & Avg. \\
    \midrule
    VDGP  & 76.3  & 49.3  & 67.8  & 17.4  & 81.5  & 66.6  & 59.8 \\
    VDGP+Ours & 77.2(+0.9) & 50.3(+1.0) & 68.3(+0.5) & 19.1(+1.7) & 82.8(+1.3) & 67.4(+0.8) & 60.9(+1.1) \\
    \midrule
    L2C   & 75.6  & 52.1  & 69.4  & 17.3  & 85.5  & 67.1  & 61.2 \\
    L2C+Ours & 76.7(+1.1) & 54.5(+2.4) & 70.1(+0.7) & 18.2(+0.9) & 86.3(+0.8) & 68.6(+1.5) & 62.4(+1.2) \\
    \bottomrule
    \bottomrule
    \end{tabular}}%
  \label{tab:test-time domain adaptation}
\end{table*}%

\begin{table*}[t]
  \centering
  \caption{Comparison results with the L2P~\cite{wang2022learning} approach on several benchmark datasets.}
  \scalebox{0.5}{
    \begin{tabular}{c|ccc|ccc|ccc|ccc|ccc}
    \toprule
    \toprule
    \multicolumn{1}{l|}{\multirow{2}[4]{*}{Method}} & \multicolumn{3}{c|}{Flower102} & \multicolumn{3}{c|}{DTD} & \multicolumn{3}{c|}{Pets} & \multicolumn{3}{c|}{UCF101} & \multicolumn{3}{c}{EuroSAT} \\
\cmidrule{2-16}          & Base  & Novel & HM    & Base  & Novel & HM    & Base  & Novel & HM    & Base  & Novel & HM    & Base  & Novel & HM \\
    \midrule
    L2P~\cite{wang2022learning}   & 31.2  & 44.0    & 36.5  & 33.2  & 33.2  & 33.2 & 63.8  & 67.4  & 65.6  & 28.3  & 25.3  & 26.7 & 34.1  & 26.9  & 30.1 \\
    Ours  & 84.2 & 71.6 & 77.4\small{(+40.9)} & 57.2 & 57.0 & 57.1\small{(+23.9)} & 94.2 & 98.0 & 96.1\small{(+30.5)} & 77.3 & 77.2 & 77.2\small{(+50.5)} & 72.7 & 70.3 & 71.5\small{(+41.4)} \\
    \bottomrule
    \bottomrule
    \end{tabular}}
  \label{tab:compare l2p}
\end{table*}%

\subsubsection{Comparison with L2P} 

We also compare our CAKI with L2P~\cite{wang2022learning} by directly using a shared prompt pool to model class-shared and class-specific knowledge. As shown in Table~\ref{tab:compare l2p}, the performance of L2P~\cite{wang2022learning} is largely inferior to our CAKI with class-specific prompts. For example, our CAKI achieves an improvement in HM accuracy over L2P~\cite{wang2022learning} by 30.5\% and 50.5\% on Pets and UCF101 datasets, respectively.
These improvements further validate the effectiveness of CAKI in modeling class-specific knowledge and injecting it into pre-trained models through prompt learning.

\subsection{Transferability to Segmentation and Detection Tasks}

\textbf{Semantic Segmentation.} To evaluate the generalizability of our approach to other vision tasks, we benchmark CAKI on four semantic segmentation datasets and their corresponding variants. For comparison, we reproduce the CoOp and CoCoOp baselines under the same segmentation settings. As shown in Table~\ref{tab:segmentation}, CAKI consistently outperforms both baselines, largely owing to its strengthened capability in semantic knowledge organization and retrieval. For example, using SEEM-Tiny, CAKI improves upon CoOp and CoCoOp by 1.0\% and 1.3\% on the BDD dataset, and by 1.9\% and 1.4\% on the ADE dataset, respectively. These results demonstrate that CAKI generalizes effectively to the semantic segmentation task.

\textbf{Object Detection.} We further assess CAKI on several object detection benchmarks. Following the same evaluation protocol, we reproduce CoOp and CoCoOp on the detection task for a direct comparison. As shown in Table~\ref{tab:detection}, CAKI achieves superior detection accuracy and robustly handles diverse conditions, including challenging weather variations. The performance gains can be attributed to CAKI’s class-aware knowledge banks, which facilitate the generation of more accurate predictions across diverse test scenarios.

\begin{table*}[t]
  \centering
  \caption{Comparison with existing methods on semantic segmentation tasks. mIoU is reported.}
  \scalebox{0.57}{
    \begin{tabular}{lccccccccc}
    \toprule
    Method & Cityscapes & BDD   & Mapillary & ADE   & Pascal & ACDC$_{Fog}$ & ACDC$_{Night}$ & ACDC$_{Rain}$ & ACDC$_{Snow}$ \\
    \midrule
    SEEM-Tiny & 39.2  & 37.4  & 42.1  & 14.6  & 45.1  & 34.6  & 20.7  & 33.1  & 35.8 \\
    CoOp  & 50.1  & 41.6  & 43.3  & 17.6  & 45.9  & 36.2  & 22.5  & 34.2  & 37.1 \\
    CoOp+Ours & 52.3(+2.2) & 42.6(+1.0) & 43.8(+0.5) & 19.5(+1.9) & 46.7(+0.8) & 37.1(+0.9) & 23.9(+1.4) & 35.6(+1.4) & 38.4(+1.3) \\
    CoCoOp & 51.2  & 38.3  & 44.5  & 20.2  & 46.6  & 36.0    & 23.7  & 35.2  & 38.2 \\
    CoCoOp+Ours & 53.3(+2.1) & 39.6(+1.3) & 44.9(+0.4) & 21.6(+1.4) & 47.1(+0.5) & 36.5(+0.5) & 25.1(+1.4) & 36.3(+1.1) & 38.5(+0.3) \\
    \midrule
    SEEM-Large & 49.3  & 44.6  & 47.9  & 15.2  & 37.1  & 48.1  & 32.0    & 47.4  & 45 \\
    CoOp  & 51.2  & 45.2  & 52.0    & 18.1  & 47.4  & 49.2  & 32.3  & 48.1  & 47.4 \\
    CoOp+Ours & 51.9(+0.7) & 46.3(+1.1) & 53.4(+1.4) & 19.3(+1.2) & 48.2(+0.8) & 50.4(+1.2) & 33.3(+1.0) & 48.7(+0.6) & 48.1(+0.7) \\
    CoCoOp & 57.1  & 49.5  & 56.2  & 25.6  & 55.3  & 49.7  & 34.3  & 47.7  & 47.6 \\
    CoCoOp+Ours & 58.6(+1.5) & 50.7(+1.2) & 57.5(+1.3) & 26.4(+0.8) & 56.8(+1.5) & 51.2(+1.5) & 34.9(+0.6) & 50.4(+2.7) & 48.5(+0.9) \\
    \bottomrule
    \end{tabular}}%
  \label{tab:segmentation}%
\end{table*}%

\begin{table*}[t]
  \centering
  \caption{Comparison with existing methods on object detection tasks. mAP$_{50}$ is reported.}
  \scalebox{0.7}{
    \begin{tabular}{lccccccc}
    \toprule
    Method & Cityscapes & BDD   & Mapillary & ACDC$_{Fog}$ & ACDC$_{Night}$ & ACDC$_{Rain}$ & ACDC$_{Snow}$ \\
    \midrule
    SEEM-Tiny & 30.5  & 26.1  & 15.7  & 44.2  & 22.3  & 25.9  & 33.9 \\
    CoOp  & 34.8  & 30.7  & 17.6  & 47.1  & 25.2  & 26.7  & 34.8 \\
    CoOp+Ours & 35.1(+0.3) & 32.3(+1.6) & 18.8(+1.2) & 47.7(+1.6) & 26.5(+1.3) & 27.4(+0.7) & 35.3(+0.5) \\
    CoCoOp & 33.7  & 31.6  & 19.4  & 46.3  & 26.7  & 26.4  & 34.5 \\
    CoCoOp+Ours & 34.6(+0.9) & 33.2(+1.6) & 20.6(+1.2) & 47.5(+1.2) & 27.1(+0.4) & 28.2(+1.8) & 36.1(+1.6) \\
    \midrule
    SEEM-Large & 31.4  & 31.8  & 18.3  & 55.2  & 31.4  & 34.8  & 43.7 \\
    CoOp  & 36.5  & 32.5  & 18.7  & 57.1  & 33.5  & 36.2  & 45.1 \\
    CoOp+Ours & 37.6(+1.1) & 33.3(+0.8) & 20.5(+1.8) & 58.4(+1.3) & 34.6(+1.1) & 36.6(+0.4) & 46.8(+1.7) \\
    CoCoOp & 37.1  & 31.9  & 19.3  & 56.3  & 33.9  & 37.7  & 44.3 \\
    CoCoOp+Ours & 38.8(+1.7) & 32.7(+0.8) & 20.9(+1.6) & 57.4(+1.1) & 34.2(+0.3) & 38.3(+0.6) & 44.8(+0.5) \\
    \bottomrule
    \end{tabular}}
  \label{tab:detection}
\end{table*}

\begin{table*}[t]
  \centering
  \caption{Ablation study of CSPG and QKPM on five base-to-novel benchmarks.}
  \scalebox{0.62}{
    \begin{tabular}{cc|ccc|ccc|ccc|ccc|ccc}
    \toprule
    \multirow{2}[4]{*}{CSPG} & \multirow{2}[4]{*}{QKPM} & \multicolumn{3}{c|}{Flower102} & \multicolumn{3}{c|}{DTD} & \multicolumn{3}{c|}{Pets} & \multicolumn{3}{c|}{UCF101} & \multicolumn{3}{c}{EuroSAT} \\
\cmidrule{3-17}          &       & Base  & Novel & HM    & Base  & Novel & HM    & Base  & Novel & HM    & Base  & Novel & HM    & Base  & Novel & HM \\
    \midrule
    \checkmark     &       & 81.1  & 71.3  & 75.8  & 55.9  & 56.3  & 56.1  & 93.1  & 97.4  & 95.2  & 76.8  & 75.9  & 76.4  & 71.5  & 69.7  & 70.6 \\
          & \checkmark     & 82.0    & 67.4  & 74.0    & 56.8  & 56.4  & 56.6  & 93.6  & 97.3  & 95.4  & 76.6  & 74.6  & 75.6  & 70.5  & 69.0    & 69.7 \\
    \checkmark     & \checkmark     & 84.2  & 71.6  & 77.4  & 57.2  & 57.0    & 57.1  & 94.2  & 98.0    & 96.1  & 77.3  & 77.2  & 77.2  & 72.7  & 70.3  & 71.5 \\
    \bottomrule
    \end{tabular}}%
  \label{tab:ablation_componet}%
\end{table*}

\subsection{Analysis and Ablation Study}

\begin{table*}[t]
  \centering
  \caption{Ablation on the effectiveness of class-shared, class-specific, and instance-specific prompts in our CAKI approach.}
    \scalebox{0.5}{
    \begin{tabular}{ccc|ccc|ccc|ccc|ccc|ccc}
    \toprule
    \toprule
    \multicolumn{1}{l}{\multirow{2}[4]{*}{Class-shared}} & \multicolumn{1}{l}{\multirow{2}[4]{*}{Instance-specific }} & \multicolumn{1}{l|}{\multirow{2}[4]{*}{Class-specific }} & \multicolumn{3}{c|}{Flower102} & \multicolumn{3}{c|}{DTD} & \multicolumn{3}{c|}{Pets} & \multicolumn{3}{c|}{UCF101} & \multicolumn{3}{c}{EuroSAT} \\
\cmidrule{4-18}          &       &       & Base  & Novel & \multicolumn{1}{c|}{HM} & Base  & Novel & \multicolumn{1}{c|}{HM} & Base  & Novel & \multicolumn{1}{c|}{HM} & Base  & Novel & HM    & Base  & Novel & HM \\
    \midrule
    \checkmark      & \checkmark      &       & 78.1  & 65.7  & 71.4  & 56.4  & 55.6  & 56.0    & 93.7  & 94.9  & 94.3  & 75.4  & 73.7  & 74.5  & 73.2  & 64.5  & 68.6 \\
    \checkmark      &       &       & 82.0    & 67.4  & \multicolumn{1}{c|}{74.0} & 56.8  & 56.4  & \multicolumn{1}{c|}{56.6} & 93.6  & 97.3  & \multicolumn{1}{c|}{95.4} & 76.6  & 74.6  & 75.6  & 70.5  & 69.0  & 69.7 \\
          & \checkmark      &       & 68.7  & 77.5  & \multicolumn{1}{c|}{72.8} & 52.9  & 60.3  & \multicolumn{1}{c|}{56.4} & 89.2  & 95.1  & \multicolumn{1}{c|}{92.1} & 70.0    & 75.8  & 72.8  & 50.1  & 65.1  & 56.6 \\
          &       & \checkmark      & 72.0    & 77.6  & \multicolumn{1}{c|}{74.7} & 56.4  & 61.6  & \multicolumn{1}{c|}{58.9} & 90.1  & 96.9  & \multicolumn{1}{c|}{93.4} & 70.1  & 75.7  & 72.8  & 52.5  & 69.9  & 60.0 \\
    \checkmark      &       & \checkmark     & 84.2  & 71.6  & \multicolumn{1}{c|}{77.4 } & 57.2  & 57.0  & \multicolumn{1}{c|}{57.1 } & 94.2  & 98.0  & \multicolumn{1}{c|}{96.1 } & 77.3  & 77.2  & 77.2  & 72.7  & 70.3  & 71.5 \\
          & \checkmark      & \checkmark      & 71.1  & 78.4  & 74.6  & 56.4  & 63.2  & 59.6  & 89.3  & 96.5  & 92.8  & 71.1  & 76.6  & 73.8  & 51.2  & 68.5  & 58.6 \\
    \bottomrule
    \bottomrule
    \end{tabular}}
  \label{tab:component_detail}
\end{table*}%

\textbf{Component-wise ablation of CSPG and QKPM.} To assess the effectiveness of each component within our framework, we conduct an ablation study using CoOp as the baseline. As shown in Table~\ref{tab:ablation_componet}, when using CSPG alone, the model is able to generate class-specific prompts but lacks the capability to identify semantically related classes. Consequently, inference relies on randomly selecting $K$ class-specific prompts, which results in a substantial degradation in performance compared with the full model. Conversely, when only QKPM is used, CAKI degenerates into relying solely on class-shared prompts, and the absence of class-level knowledge leads to a notable reduction in accuracy.
By combining both components, CAKI can retrieve semantically relevant category knowledge and generate tailored prompts for different test samples. This enables CAKI to achieve the strong performance across multiple benchmark datasets. These results indicate that CSPG and QKPM provide complementary benefits and that both components are essential for the effectiveness of CAKI.

\textbf{Class-shared and instance-specific prompts \textit{vs.} class-specific prompts.} 
To better understand the characteristics of learned class-specific knowledge, we compare class-shared and instance-specific prompts with class-specific prompts and present the accuracies for several benchmark datasets in Table~\ref{tab:component_detail}. It can be observed that using class-shared and instance-specific prompts separately is more effective than using both simultaneously. This is primarily because shared knowledge across classes significantly reduces the recognition accuracy of novel classes in instance-specific prompts, indicating a conflict between two types of knowledge. Different from this, our class-specific prompts are complementary and can consistently improve the performance of class-shared and instance-specific prompts by injecting class-level knowledge into pre-trained models.

\begin{table*}[t]
  \centering
  \caption{Effectiveness analysis of prompt matching strategy with CoOp~\cite{zhou2022learning} approach. Prompt-A: all prompts ensemble; Prompt-R: random prompts ensemble; Prompt-M: query-key prompt matching.}
  \scalebox{0.49}{
    \begin{tabular}{c|ccc|ccc|cccccc|ccc}
    \toprule
    \toprule
    \multicolumn{1}{c|}{\multirow{2}[4]{*}{Method}} & \multicolumn{3}{c|}{Flower102} & \multicolumn{3}{c|}{DTD} & \multicolumn{3}{c|}{Pets} & \multicolumn{3}{c|}{UCF101} & \multicolumn{3}{c}{EuroSAT} \\
\cmidrule{2-16}          & Base  & Novel & HM    & Base  & Novel & HM    & \multicolumn{1}{c}{Base} & Novel & \multicolumn{1}{c|}{HM} & Base  & Novel & HM    & Base  & Novel & HM \\
    \midrule
    Prompt-A & 84.1  & 68.7  & 75.6  & 56.2  & 55.6  & 55.9  & \multicolumn{1}{c}{94.1} & 97.4  & \multicolumn{1}{c|}{95.7} & 76.8  & 75.7  & 76.3  & 71.6  & 69.8  & 70.6 \\
    Prompt-R & 81.1  & 71.3  & 75.8  & 55.9  & 56.3  & 56.1  & \multicolumn{1}{c}{93.1} & 97.4  & \multicolumn{1}{c|}{95.2} & 76.8  & 75.9  & 76.4  & 71.5  & 69.7  & 70.6 \\
    Prompt-M (Ours)  & \textbf{84.2}  & \textbf{71.6}  & \textbf{77.4}  & \textbf{57.2}  & \textbf{57.0}  & \textbf{57.1}  & \textbf{94.2}  & \textbf{98.0}  & \textbf{96.1}  & \textbf{77.3}  & \textbf{77.2}  & \textbf{77.2}  & \textbf{72.7}  & \textbf{70.3}  & \textbf{71.5} \\
    \bottomrule
    \bottomrule
    \end{tabular}}
  \label{tab:promptmatching}
\end{table*}%

\begin{figure}[t]
  \centering
  \includegraphics[width=0.6\textwidth]{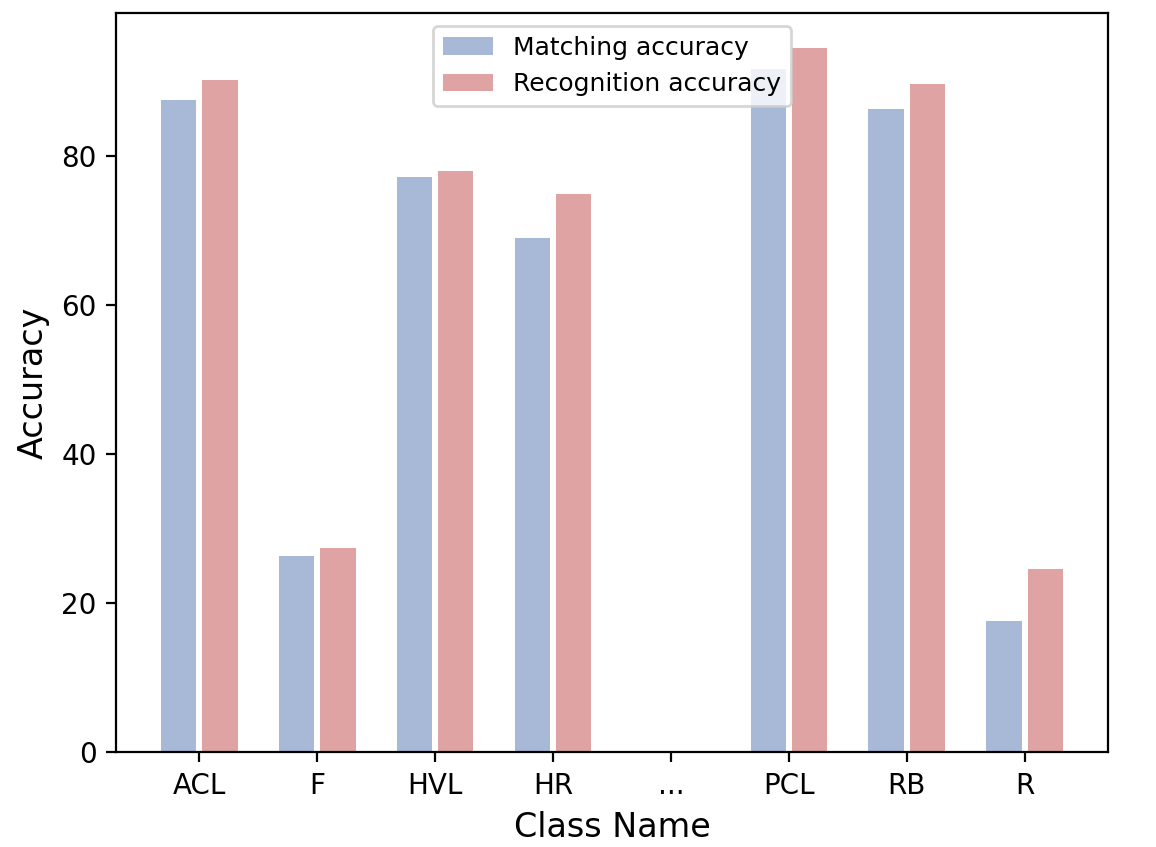}\vspace{-3pt}
  \caption{Comparison between recognition accuracy and the matching accuracy of CAKI, using EuroSAT dataset. ACL: Annual Crop Land; F: Forest; HVL: Herbaceous Vegetation Land; HR: Highway or Road; PCL: Permanent Crop Land; RB: Residential Buildings; R: River.
  }\label{fig:match}\vspace{-14pt}
\end{figure}

\par\textbf{Effectiveness of prompt matching.} In Table~\ref{tab:promptmatching}, we conduct an analysis of the efficacy of the prompt matching strategy with the CoOp approach. We explore three kinds of matching strategies, wherein ``Prompt-R" denotes the randomly selected prompt ensemble, ``Prompt-A" indicates the aggregation of all prompts, and ``Prompt-M" represents our proposed query-key prompt matching technique. As shown in Table~\ref{tab:promptmatching}, our method outperforms the two alternative strategies in terms of overall performance on downstream tasks. With the query-key matching mechanism (Prompt-M), CAKI instructs the model to explore class-specific knowledge and dynamically align with the test sample. In contrast, other two strategies either fail to select useful prompts (Prompt-R) or suffer from a negative impact of redundant information (Prompt-A). This experimentally demonstrated that the pre-trained models benefit from the specific design of our approach motivated by the fact that test samples belonging to different classes often exhibit significant similarities in certain visual patterns. For example, CAKI with Prompt-M achieves an accuracy of 57.2\% on the DTD dataset, while Prompt-R and Prompt-A only reach 56.2\% and 55.9\%, respectively, on the same dataset. This trend is similarly observed across other datasets, indicating that our proposed matching strategy is robust across different downstream tasks.

\textbf{Relationship between class-specific and class-shared prompts.}
Our method utilizes class-shared prompts to generate accurate matching scores while modeling prior knowledge about the class to provide the model with class-specific prompts. In this experiment, we explore the relationship between class-specific and class-shared prompts using the EuroSAT dataset. Figure~\ref{fig:match} reports the matching accuracy and recognition accuracy of our CAKI for each class. As shown in Figure~\ref{fig:match}, our method achieves higher accuracy in most categories compared to matching accuracy. Notably, the improvement is most pronounced in the Highway or Road (HR) class. This is primarily because, in other scenarios, there are no scenes similar to highways, leading to underfitting with class-shared prompts. Our method effectively addresses this issue by providing class-level knowledge injection, thereby improving the model performance on this class.

\begin{figure}[t]
  \centering
  \includegraphics[width=0.75\textwidth]{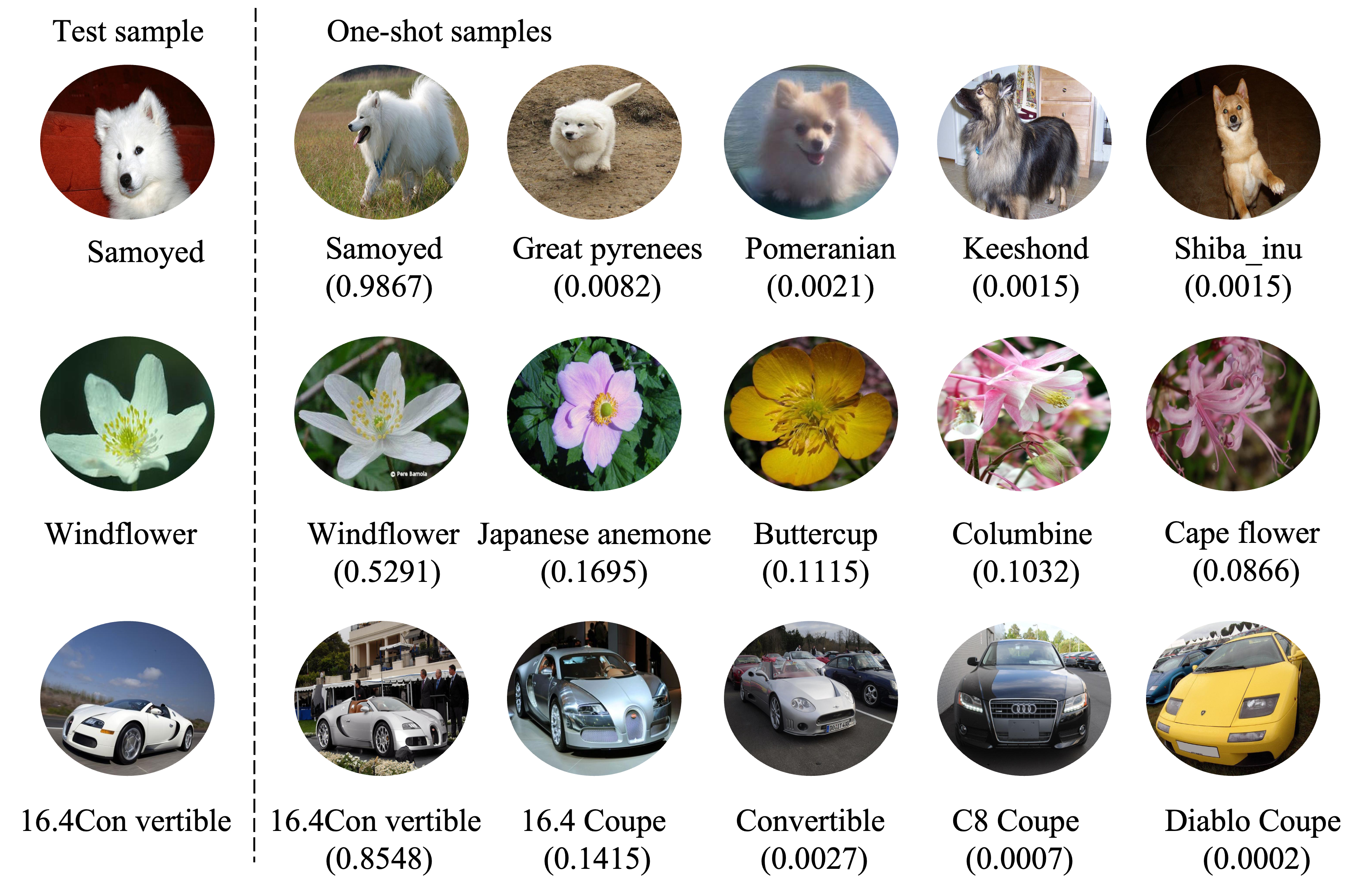}
  \caption{The visualization of matched samples and their labels (matching scores) with respective to the test sample, using Pets, Flower102, and Cars datasets.
  }\label{fig:visualisation}\vspace{-3pt}
\end{figure}

\textbf{CAKI visualization.} In Figure~\ref{fig:visualisation}, we showcase the visualization of labeled samples projected into class-aware prompts, using three different datasets: Pets, Flower102, and Cars.  CAKI can automatically match test samples with the class that shares similar visual patterns, demonstrating the model's capability to retrieve semantically similar prompts from key-value memory. Moreover, 
we find that the model matches test images with similar visual patterns but with different degrees of confidence. Benefiting from CAKI, our method consistently exhibits high confidence in prompts belonging to the same class as the test samples.

\subsection{Scalability of our CAKI} 

Our class-specific prompts do not acquire additional GPU memory during training or inference. Instead, we store these prompts locally, which incurs additional parameter storage requirements. As shown in Table~\ref{tab:scalability}, the learned parameter (prompts) storage and training (testing) time of our CAKI exhibits a linear growth trend with the number of classes on Flower102 dataset. This demonstrates that our approach maintains acceptable storage and computational overhead, ensuring scalability even with a significant increase in classes.
\begin{table*}[t]
       \begin{center}
    \caption{Evaluation results of our CAKI on the learned parameter (prompts) storage and training (testing) time.}
       \scalebox{0.85}{
    \setlength{\tabcolsep}{0.08cm}{
    \begin{tabular}{lccccc}
    \hline
    \hline
    No. of class & 20    & 40    & 61    & 81    & 102 \\
    \hline
    No. of prompt & 20    & 40    & 61    & 81    & 102 \\
    \hline
    Train (min) & 0.32 & 0.55 & 1.06 & 1.20 & 1.50 \\
    Test (min) & 1.20 & 2.32 & 4.24 & 6.14 & 7.82 \\
    Storage (G) & 0.32  & 0.64  & 0.96  & 1.28  & 1.60 \\
    \hline
    \hline
    \end{tabular}}}\label{tab:scalability}
      \end{center}
      \vspace{-3mm}
\end{table*}%

\begin{table}[t]
  \centering
  \caption{Computational cost of CAKI for base-to-novel generalization in 1-shot setting. Per-image inference time and harmonic mean is reported for different values of $K$. We adopt CoOp as coarse model.}
    \begin{tabular}{cccccc}
    \toprule
    \multirow{2}[4]{*}{Method} & \multirow{2}[4]{*}{CoOp} & \multirow{2}[4]{*}{CoCoOp} &       & CAKI(Ours)  &  \\
\cmidrule{4-6}          &       &       & K=1   & K=3   & K=5 \\
    \midrule
    Inference time (s/img) & 0.064 & 0.067 & 0.092 & 0.146 & 0.196 \\
    Harmonic mean    & 74.0  & 75.0  & 77.1  & 77.4  & 76.9 \\
    \bottomrule
    \end{tabular}
  \label{tab:per-img inference}
\end{table}

Table~\ref{tab:per-img inference} shows the per-image inference time of CAKI under different values of $K$. As 
$K$ increases, inference latency grows due to the retrieval and aggregation of more class-specific prompts. With $K=3$, our CAKI achieves the significant improvement of +3.3\% with an acceptable inference time of 0.146s/img. These results demonstrate that CAKI effectively balances accuracy and efficiency and is suitable for practical deployment.

\subsection{The Effect of Prompt Cache Size}

We reduce the size of the prompt cache to $\frac{1}{2}$/$\frac{1}{4}$ of its original size using the following methods: (1) randomly select $\frac{1}{2}$/$\frac{1}{4}$ of classes and only learn prompts for the selected classes; or (2) enforce each 2/4 similar classes to \textit{share/use} the same prompt. The results in Table~\ref{tab:prompt cache} show that while our method slightly decreases as cache size reduces, it still consistently outperforms existing methods due to its automatic incorporation of class-relevant context knowledge instead of relying on remembering fixed classes. This further demonstrates the effectiveness of our knowledge injection.
\begin{table*}[t]
       \begin{center}
        \caption{Evaluation results of our CAKI under different prompt cache sizes.}
       \scalebox{0.7}{
        \setlength{\tabcolsep}{0.08cm}{
    \begin{tabular}{l|cc|cc|c}
    \toprule
    Prompt size  & \multicolumn{2}{c|}{20} & \multicolumn{2}{c|}{61} & \multicolumn{1}{c}{102} \\
    \midrule
    Method & Randomly & Share & Randomly & Share & Original \\
    \midrule
    CoOp  & 94.2  & 94.5  & 94.8  & 95.0    & 95.0   \\
    CAKI (Ours) & \textbf{94.3}  & \textbf{94.9}  & \textbf{95.1}  & \textbf{95.4 }   & \textbf{95.7}  \\
    \bottomrule
    \end{tabular}}}\label{tab:prompt cache}
      \end{center}
      \vspace{-8mm}
\end{table*}

\begin{figure*}[t]
  \centering
  \includegraphics[width=0.9\textwidth]{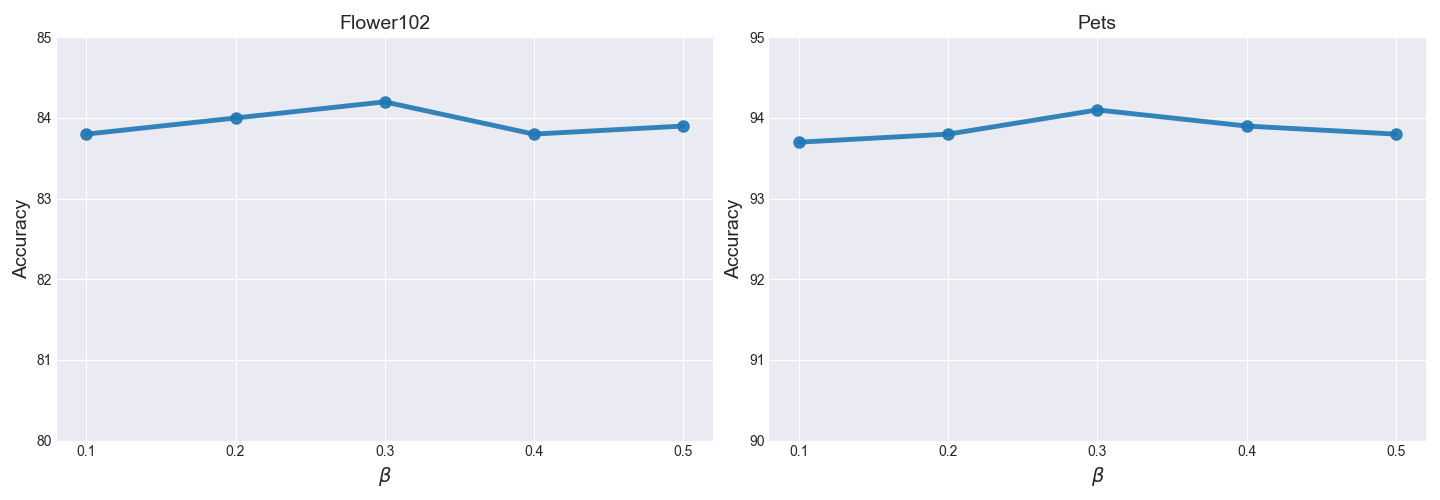}
  \caption{Parameter study on the weight parameter $\beta$ that balances coarser and finer predictions, conducted using the Flower102 and Pets datasets under the 1-shot base-to-novel setting. Classification accuracy on base classes is reported.
  }\label{fig:beta}
\end{figure*}

\begin{figure*}[t]
  \centering
  \includegraphics[width=0.9\textwidth]{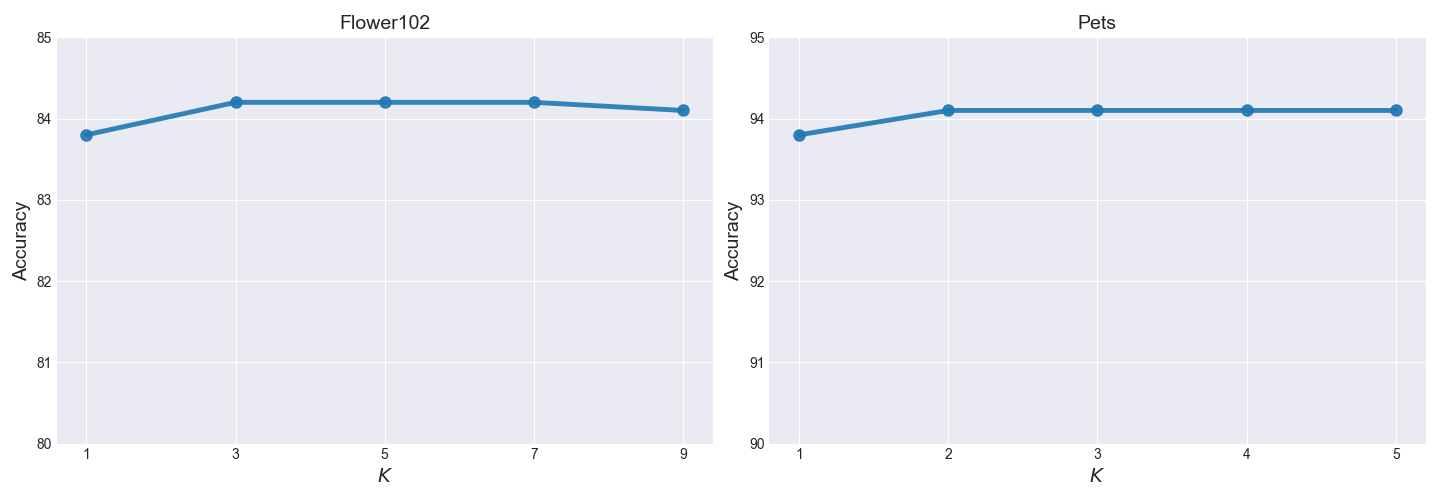}
  \caption{Parameter studies on the varying number of top-$K$ class-specific prompts retrieved from key-value memory, conducted using the Flower102 and Pets datasets under the 1-shot base-to-novel setting. Classification accuracy on base classes is reported.
  }\label{fig:topk}
\end{figure*}

\subsection{Hyperparameter Analysis} 

We first investigate the impact of the weight parameter 
$\beta$ on our method under base-to-novel setting, and report the results for the Flower102 and Pets datasets, as presented in Figure~\ref{fig:beta}. Experimental results on the Flower102 and Pets datasets demonstrate that our approach achieves obvious improvements as $\beta$ increases, attaining optimal performance when $\beta$ is set to 0.3. However, a decline in the model's performance appears when $\beta$ continuously increases. When $\beta$ is too large, the model overweights class-specific prompts that are learned from limited samples, leading to degraded generalization. When $\beta$ is too small, the discriminative information encoded in class-specific prompts is underutilized, limiting the model's ability to distinguish between similar categories. As shown in Figure~\ref{fig:beta}, $\beta = 0.3$ yields the best performance across both datasets, reflecting an effective balance between these two complementary sources of knowledge.


We also conduct a sensitivity evaluation of the hyperparameter $K$ on Flower102 and Pets datasets. We kept other components fixed and ran our CAKI method with $K=1, 3, 5, 7, 9$, respectively. As shown in Figure~\ref{fig:topk}, increasing $K$ allows the model to retrieve more relevant semantic knowledge, improving performance; however, when $K$ becomes too large (e.g., $K=9$), the retrieved set inevitably includes semantically irrelevant or noisy classes, which dilutes the discriminative cues and leads to the performance drop. To avoid the repeated dataset-specific tuning, we set $K=3$ as the default across all datasets.

\begin{table}[t]
  \centering
  \caption{The effect of temperature parameter $\tau$ on Flower102 and Pets datasets.}\vspace{-3mm}
    \begin{tabular}{lccccc}
    \toprule
    Temperature $\tau$ & 0.6   & 0.8   & 1.0     & 1.2   & 1.4 \\
    \midrule
    Flower102 & 83.8  & 84.0    & 84.2  & 84.1  & 84.0  \\
    Pets  & 93.9  & 94.1  & 94.2  & 94.0     & 94.1 \\
    \bottomrule
    \end{tabular}
  \label{tab:temperature parameter}
\end{table}

\begin{figure*}[t]
  \centering
  \includegraphics[width=0.8\textwidth]{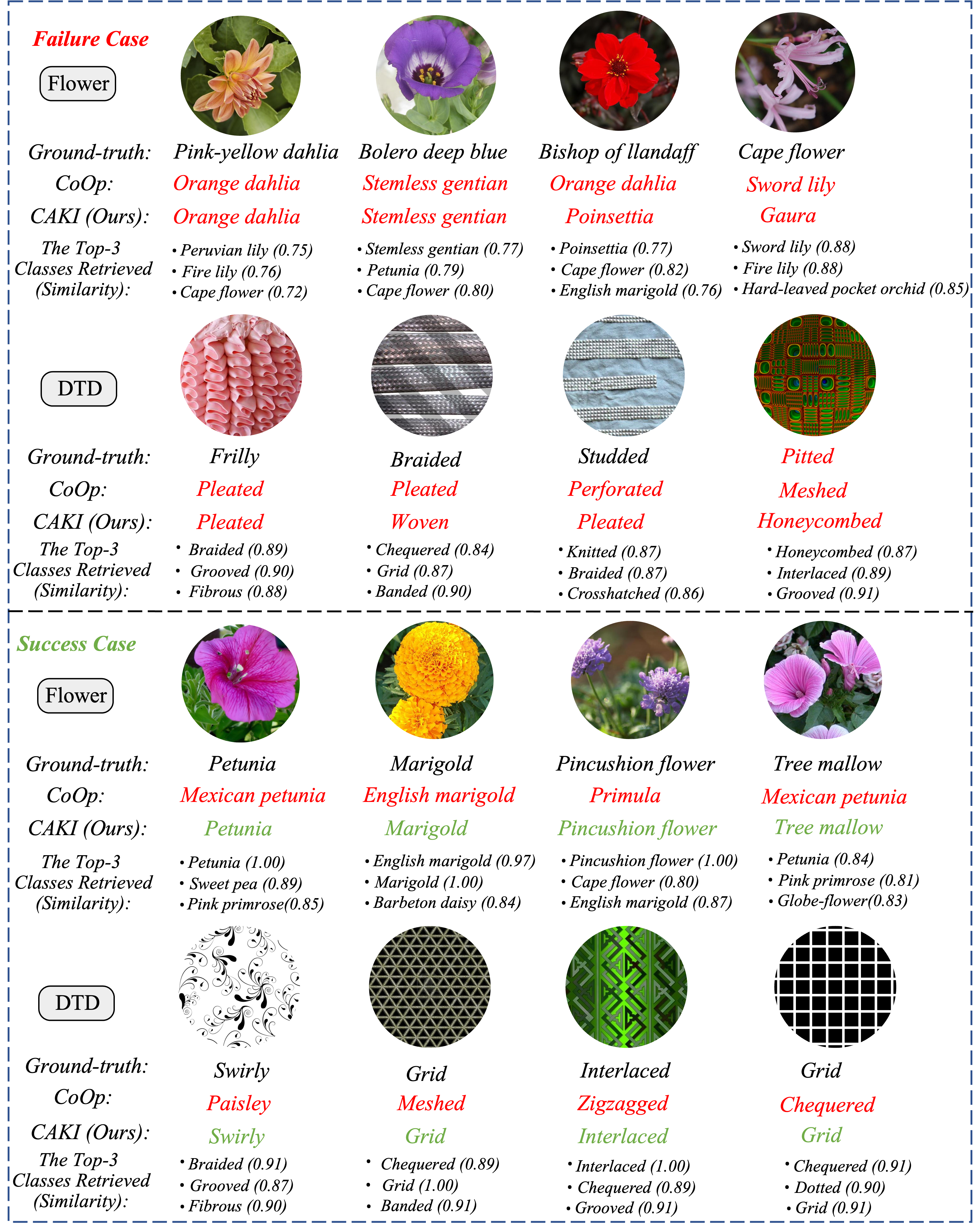}
  \caption{Qualitative failure and success cases on Flower (Flower-102) and Texture (DTD) datasets. Top: failure cases where both the coarse model and our method misclassify, revealing challenges in resolving semantic ambiguity. Bottom: success cases where our method corrects coarse-model errors, demonstrating the efficacy of class-level knowledge retrieval in refining predictions. To further quantify semantic relevance, we report the similarity between the retrieved semantic classes and the ground-truth labels using the CLIP text encoder.
  }\label{fig:failure}
\end{figure*}

\begin{figure*}[t]
  \begin{center}
    \includegraphics[width=0.9\textwidth]{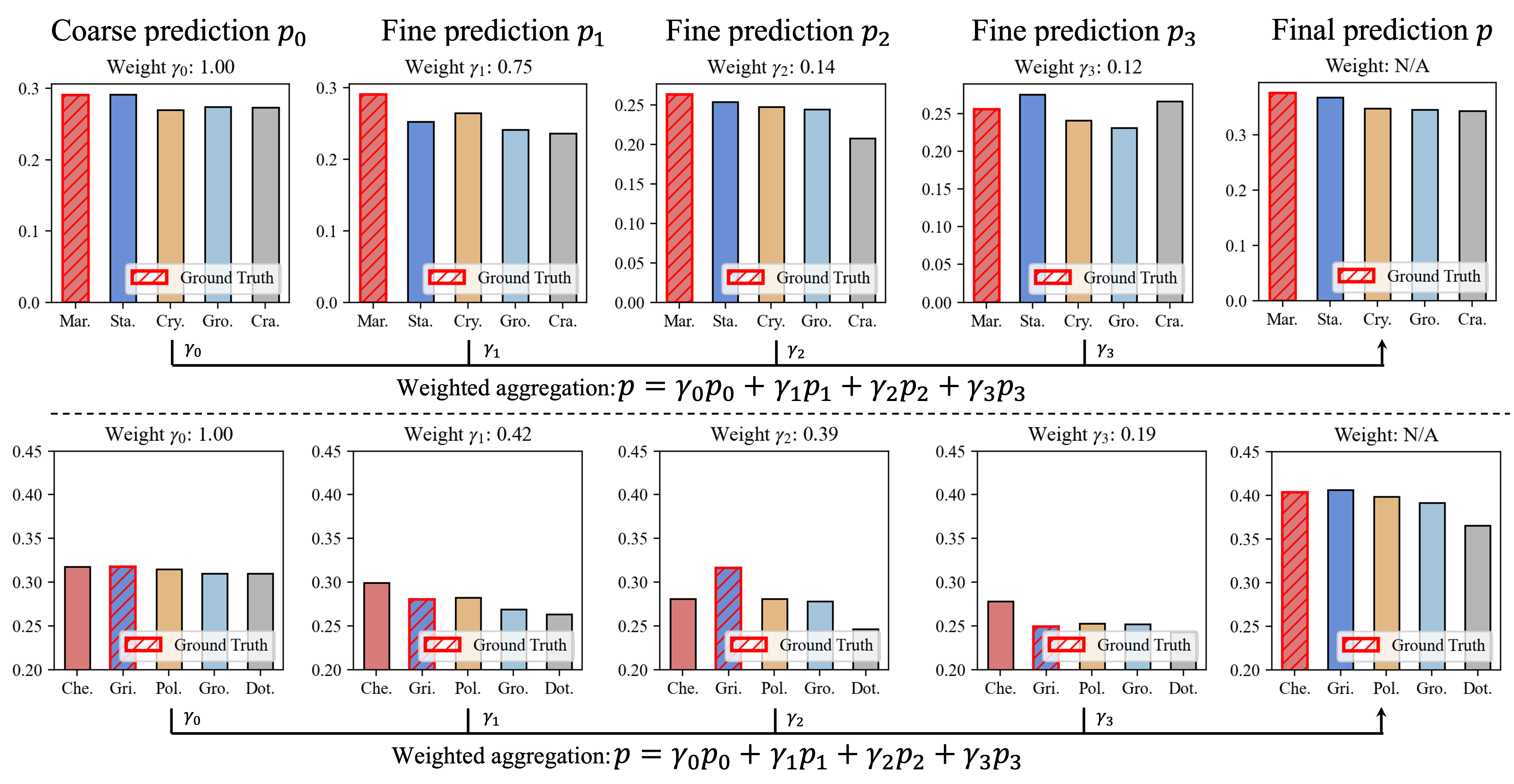}\vspace{-2mm}
    \caption{Visualization of the initial prediction, the fine predictions with class-specific prompts, and their corresponding weights—i.e., the matching scores obtained from the coarse model’s responses to different categories—illustrating how CAKI corrects the initial misprediction by leveraging discriminative information among semantically similar classes. Mar. denotes marbled, Sta. denotes stained, Cry. denotes crystalline, Gro. denotes grooved, Cra. denotes cracked, Gri. denotes grid, Che. denotes chequered, Pol. denotes polka-dotted, Dot. denotes dotted, and Per. denotes perforated.}\label{fig:rectification}
  \end{center}
\vspace{-1.8em}
\end{figure*}

We further investigate the effect of the temperature parameter $\tau$, which controls the sharpness of the probability distribution. As shown in Table~\ref{tab:temperature parameter}, the performance variation of CAKI across a wide range of $\tau$ values $[0.6, 1.4]$ is within 0.5\%, indicating that our method is generally robust to the choice of $\tau$. These results suggest that the temperature parameter acts primarily as a scaling factor rather than a critical hyperparameter.

Figure~\ref{fig:failure} presents qualitative examples on the Flower and DTD datasets, illustrating both failure and success cases of CAKI in comparison with CoOp. As shown in Figure~\ref{fig:failure}, errors typically arise when visual ambiguity among fine-grained classes is high, leading CAKI to retrieve class-level knowledge that lacks semantic relevance. For example, the retrieved semantic classes in failure cases consistently exhibit lower CLIP-based similarity to the ground-truth label than those in success cases. On the other hand, our method also tends to fail on certain novel categories, such as ``Pink-yellow dahlia" in the Flower dataset and ``Pitted" in the DTD dataset. The primary reason may be that novel class exhibits semantic characteristics that are not well covered by base classes. These cases illustrate the conditions in which CAKI is less effective and provide directions for improvement, such as LLM-based semantic knowledge expansion.

In contrast, the success cases (bottom) highlight scenarios in which CAKI effectively leverages class-level knowledge to correct mispredictions made by CoOp. By retrieving class-relevant prompts, CAKI produces more discriminative representations, enabling it to distinguish visually similar flower categories (e.g., Mexican petunia vs. Petunia) and texture types (e.g., Meshed vs. Grid). These examples demonstrate how class-level knowledge contributes to CAKI’s improved robustness in challenging fine-grained settings. Figure~\ref{fig:rectification} also shows that CAKI effectively rectifies the inaccurate predictions of coarse model, particularly under high-uncertainty scenarios. For example, although coarse model incorrectly predicts the test sample as belonging to the ``Stained” class with high uncertainty, injecting class-level knowledge enables the model to acquire class-specific semantic cues and subsequently infer the correct category (i.e., ``Marbled”). These results demonstrate that CAKI effectively aggregates the useful information from class-level knowledge stored in class-specific prompts, thereby mitigating biases inherited from the coarse model.

\section{Conclusion}

In this paper, we introduced the CAKI framework, a novel approach for enhancing vision-language models (VLMs) by integrating class-specific knowledge into prompt learning. Our method effectively addresses the limitations of existing class-shared and instance-specific prompts by injecting class-level information, thereby improving model performance on a variety of downstream tasks. Extensive experiments demonstrated that CAKI not only outperforms several existing methods but also generalizes well to novel classes.

\textbf{Limitation.} While CAKI achieves strong performance and generalizes well to novel classes, its effectiveness remains partly constrained by the coarse model’s semantic reliability. When coarse predictions lack meaningful proximity to the true class, the retrieved knowledge may become less informative. In addition, CAKI may struggle with novel classes whose semantic knowledge is not well represented by base-class knowledge. One potential remedy is to leverage large language models (LLMs) for adaptive semantic expansion, which may provide richer contextual cues for unseen or semantically distant classes. Another direction is to incorporate entropy-based uncertainty modeling to enable more reliable and robust knowledge retrieval.

\section{Data Availability}
The authors declare that
the data supporting the experiments in this study are publicly
available in the repository, https://github.com/azshue/TPT.

\section{Acknowledgments}
This work was partially supported by National Natural Science
Foundation of China (NSFC) No. 62406037, 62372150,
Postdoctoral Fellowship Program of China No. GZC20230319, Beijing
Natural Science Foundation Joint Fund Project No. L247010, China.

\bibliography{sn-bibliography}

@String(IJCV = {Int. J. Comput. Vis.})

@String(CVPR= {IEEE Conf. Comput. Vis. Pattern Recog.})

@String(ICCV= {Int. Conf. Comput. Vis.})

@String(ECCV= {Eur. Conf. Comput. Vis.})

@String(NIPS= {Adv. Neural Inform. Process. Syst.})

@String(CVPRW= {IEEE Conf. Comput. Vis. Pattern Recog. Worksh.})

@String(ICML = {Int. Conf. Mach. Learn.})

@article{merity2016pointer,
  title={Pointer sentinel mixture models},
  author={Merity, Stephen and Xiong, Caiming and Bradbury, James and Socher, Richard},
  journal={arXiv preprint arXiv:1609.07843},
  year={2016}
}

@inproceedings{radford2021learning,
  title={Learning transferable visual models from natural language supervision},
  author={Radford, Alec and Kim, Jong Wook and Hallacy, Chris and Ramesh, Aditya and Goh, Gabriel and Agarwal, Sandhini and Sastry, Girish and Askell, Amanda and Mishkin, Pamela and Clark, Jack and others},
  booktitle=ICML,
  pages={8748--8763},
  year={2021},
  organization={PMLR}
}

@article{lee2022uniclip,
  title={Uniclip: Unified framework for cocoopntrastive language-image pre-training},
  author={Lee, Janghyeon and Kim, Jongsuk and Shon, Hyounguk and Kim, Bumsoo and Kim, Seung Hwan and Lee, Honglak and Kim, Junmo},
  journal=NIPS,
  volume={35},
  pages={1008--1019},
  year={2022}
}

@inproceedings{nilsback2008automated,
  title={Automated flower classification over a large number of classes},
  author={Nilsback, Maria-Elena and Zisserman, Andrew},
  booktitle={2008 Sixth Indian Conference on Computer Vision, graphics \& image processing},
  pages={722--729},
  year={2008},
  organization={IEEE}
}

@inproceedings{cimpoi2014describing,
  title={Describing textures in the wild},
  author={Cimpoi, Mircea and Maji, Subhransu and Kokkinos, Iasonas and Mohamed, Sammy and Vedaldi, Andrea},
  booktitle=CVPR,
  pages={3606--3613},
  year={2014}
}

@inproceedings{parkhi2012cats,
  title={Cats and dogs},
  author={Parkhi, Omkar M and Vedaldi, Andrea and Zisserman, Andrew and Jawahar, CV},
  booktitle=CVPR,
  pages={3498--3505},
  year={2012},
  organization={IEEE}
}

@inproceedings{bossard2014food,
  title={Food-101--mining discriminative components with random forests},
  author={Bossard, Lukas and Guillaumin, Matthieu and Van Gool, Luc},
  booktitle=ECCV,
  pages={446--461},
  year={2014},
  organization={Springer}
}

@inproceedings{krause20133d,
  title={3d object representations for fine-grained categorization},
  author={Krause, Jonathan and Stark, Michael and Deng, Jia and Fei-Fei, Li},
  booktitle=CVPRW,
  pages={554--561},
  year={2013}
}

@inproceedings{xiao2010sun,
  title={Sun database: Large-scale scene recognition from abbey to zoo},
  author={Xiao, Jianxiong and Hays, James and Ehinger, Krista A and Oliva, Aude and Torralba, Antonio},
  booktitle=CVPR,
  pages={3485--3492},
  year={2010},
  organization={IEEE}
}

@article{helber2019eurosat,
  title={Eurosat: A novel dataset and deep learning benchmark for land use and land cover classification},
  author={Helber, Patrick and Bischke, Benjamin and Dengel, Andreas and Borth, Damian},
  journal={IEEE Journal of Selected Topics in Applied Earth Observations and Remote Sensing},
  volume={12},
  number={7},
  pages={2217--2226},
  year={2019},
  publisher={IEEE}
}

@article{soomro2012ucf101,
  title={UCF101: A dataset of 101 human actions classes from videos in the wild},
  author={Soomro, Khurram and Zamir, Amir Roshan and Shah, Mubarak},
  journal={arXiv preprint arXiv:1212.0402},
  year={2012}
}

@article{li2024prompt,
  title={PromptKD: Unsupervised Prompt Distillation for Vision-Language Models},
  author={Zheng ,Li and Xiang, Li and Xinyi, Fu and Xing, Zhang and Weiqiang, Wang, Jian and Yang},
  journal=CVPR,
  year={2024}
}

@inproceedings{fei2004learning,
  title={Learning generative visual models from few training examples: An incremental bayesian approach tested on 101 object categories},
  author={Fei-Fei, Li and Fergus, Rob and Perona, Pietro},
  booktitle=CVPR,
  pages={178--178},
  year={2004},
  organization={IEEE}
}

@article{zhou2022learning,
  title={Learning to prompt for vision-language models},
  author={Zhou, Kaiyang and Yang, Jingkang and Loy, Chen Change and Liu, Ziwei},
  journal=IJCV,
  @volume={130},
  @number={9},
  @pages={2337--2348},
  year={2022},
  @publisher={Springer}
}

@inproceedings{zhou2022conditional,
  title={Conditional prompt learning for vision-language models},
  author={Zhou, Kaiyang and Yang, Jingkang and Loy, Chen Change and Liu, Ziwei},
  booktitle=CVPR,
  @pages={16816--16825},
  year={2022}
}

@inproceedings{khattak2023maple,
  title={Maple: Multi-modal prompt learning},
  author={Khattak, Muhammad Uzair and Rasheed, Hanoona and Maaz, Muhammad and Khan, Salman and Khan, Fahad Shahbaz},
  booktitle=CVPR,
  pages={19113--19122},
  year={2023}
}

@inproceedings{prompt_nlp2,
  author       = {Brian Lester and
                  Rami Al{-}Rfou and
                  Noah Constant},
  
  title        = {The Power of Scale for Parameter-Efficient Prompt Tuning},
  booktitle    = {EMNLP},
  @pages        = {3045--3059},
  @publisher    = {Association for Computational Linguistics},
  year         = {2021},
  
}

@article{DBLP:journals/corr/abs-2111-03930,
  author       = {Renrui Zhang and
                  Rongyao Fang and
                  Wei Zhang and
                  Peng Gao and
                  Kunchang Li and
                  Jifeng Dai and
                  Yu Qiao and
                  Hongsheng Li},
  title        = {Tip-Adapter: Training-free CLIP-Adapter for Better Vision-Language
                  Modeling},
  journal      = {arXiv preprint arXiv:2111.03930},

  year         = {2021},
 
}

@inproceedings{prompt_nlp1,
  author       = {Xiang Lisa Li and
                  Percy Liang},

  title        = {Prefix-Tuning: Optimizing Continuous Prompts for Generation},
  booktitle    = {ACL},
  @pages        = {4582--4597},

  year         = {2021},
 
}

@article{bahng2022exploring,
  title={Exploring visual prompts for adapting large-scale models},
  author={Bahng, Hyojin and Jahanian, Ali and Sankaranarayanan, Swami and Isola, Phillip},
  journal={arXiv preprint arXiv:2203.17274},
  year={2022}
}

@article{shu2022test,
  title={Test-time prompt tuning for zero-shot generalization in vision-language models},
  author={Shu, Manli and Nie, Weili and Huang, De-An and Yu, Zhiding and Goldstein, Tom and Anandkumar, Anima and Xiao, Chaowei},
  journal=NIPS,
  volume={35},
  pages={14274--14289},
  year={2022}
}

@inproceedings{jia2022visual,
  title={Visual prompt tuning},
  author={Jia, Menglin and Tang, Luming and Chen, Bor-Chun and Cardie, Claire and Belongie, Serge and Hariharan, Bharath and Lim, Ser-Nam},
  booktitle=ECCV,
  pages={709--727},
  year={2022},
  organization={Springer}
}

@inproceedings{jia2021scaling,
  title={Scaling up visual and vision-language representation learning with noisy text supervision},
  author={Jia, Chao and Yang, Yinfei and Xia, Ye and Chen, Yi-Ting and Parekh, Zarana and Pham, Hieu and Le, Quoc and Sung, Yun-Hsuan and Li, Zhen and Duerig, Tom},
  booktitle=ICML,
  pages={4904--4916},
  year={2021},
  organization={PMLR}
}

@article{maji2013fine,
  title={Fine-grained visual classification of aircraft},
  author={Maji, Subhransu and Rahtu, Esa and Kannala, Juho and Blaschko, Matthew and Vedaldi, Andrea},
  journal={arXiv preprint arXiv:1306.5151},
  year={2013}
}

@article{zang2022unified,
  title={Unified vision and language prompt learning},
  author={Zang, Yuhang and Li, Wei and Zhou, Kaiyang and Huang, Chen and Loy, Chen Change},
  journal={arXiv preprint arXiv:2210.07225},
  year={2022}
}

@inproceedings{wang2022learning,
  title={Learning to prompt for continual learning},
  author={Wang, Zifeng and Zhang, Zizhao and Lee, Chen-Yu and Zhang, Han and Sun, Ruoxi and Ren, Xiaoqi and Su, Guolong and Perot, Vincent and Dy, Jennifer and Pfister, Tomas},
  booktitle=CVPR,
  pages={139--149},
  year={2022}
}

@article{abdul2024align,
  title={Align Your Prompts: Test-Time Prompting with Distribution Alignment for Zero-Shot Generalization},
  author={Abdul Samadh, Jameel and Gani, Mohammad Hanan and Hussein, Noor and Khattak, Muhammad Uzair and Naseer, Muhammad Muzammal and Shahbaz Khan, Fahad and Khan, Salman H},
  journal=NIPS,
  volume={36},
  year={2024}
}

@article{liu2023pre,
  title={Pre-train, prompt, and predict: A systematic survey of prompting methods in natural language processing},
  author={Liu, Pengfei and Yuan, Weizhe and Fu, Jinlan and Jiang, Zhengbao and Hayashi, Hiroaki and Neubig, Graham},
  journal={ACM Computing Surveys},
  volume={55},
  number={9},
  pages={1--35},
  year={2023},
  publisher={ACM New York, NY}
}

@article{gao2024clip,
  title={Clip-adapter: Better vision-language models with feature adapters},
  author={Gao, Peng and Geng, Shijie and Zhang, Renrui and Ma, Teli and Fang, Rongyao and Zhang, Yongfeng and Li, Hongsheng and Qiao, Yu},
  journal=IJCV,
  @volume={132},
  @number={2},
  @pages={581--595},
  year={2024},
  @publisher={Springer}
}

@inproceedings{feng2023diverse,
  title={Diverse data augmentation with diffusions for effective test-time prompt tuning},
  author={Feng, Chun-Mei and Yu, Kai and Liu, Yong and Khan, Salman and Zuo, Wangmeng},
  booktitle=ICCV,
  @pages={2704--2714},
  year={2023}
}

@inproceedings{karmanov2024efficient,
  title={Efficient Test-Time Adaptation of Vision-Language Models},
  author={Karmanov, Adilbek and Guan, Dayan and Lu, Shijian and El Saddik, Abdulmotaleb and Xing, Eric},
  booktitle=CVPR,
  pages={14162--14171},
  year={2024}
}

@article{grave2017unbounded,
  title={Unbounded cache model for online language modeling with open vocabulary},
  author={Grave, Edouard and Cisse, Moustapha M and Joulin, Armand},
  journal=NIPS,
  volume={30},
  year={2017}
}

@inproceedings{khandelwal2024promptsync,
  title={PromptSync: Bridging Domain Gaps in Vision-Language Models through Class-Aware Prototype Alignment and Discrimination},
  author={Khandelwal, Anant},
  booktitle=CVPR,
  pages={7819--7828},
  year={2024}
}

@inproceedings{yao2024tcp,
  title={TCP: Textual-based Class-aware Prompt tuning for Visual-Language Model},
  author={Yao, Hantao and Zhang, Rui and Xu, Changsheng},
  booktitle=CVPR,
  pages={23438--23448},
  year={2024}
}

@article{he2021towards,
  title={Towards a unified view of parameter-efficient transfer learning},
  author={He, Junxian and Zhou, Chunting and Ma, Xuezhe and Berg-Kirkpatrick, Taylor and Neubig, Graham},
  journal={arXiv preprint arXiv:2110.04366},
  year={2021}
}

@inproceedings{khattak2023self,
  title={Self-regulating prompts: Foundational model adaptation without forgetting},
  author={Khattak, Muhammad Uzair and Wasim, Syed Talal and Naseer, Muzammal and Khan, Salman and Yang, Ming-Hsuan and Khan, Fahad Shahbaz},
  booktitle=ICCV,
  pages={15190--15200},
  year={2023}
}

@article{miyai2023locoop,
  title={Locoop: Few-shot out-of-distribution detection via prompt learning},
  author={Miyai, Atsuyuki and Yu, Qing and Irie, Go and Aizawa, Kiyoharu},
  journal=NIPS,
  volume={36},
  pages={76298--76310},
  year={2023}
}

@inproceedings{lafon2024gallop,
  title={Gallop: Learning global and local prompts for vision-language models},
  author={Lafon, Marc and Ramzi, Elias and Rambour, Cl{\'e}ment and Audebert, Nicolas and Thome, Nicolas},
  booktitle=ECCV,
  pages={264--282},
  year={2024},
  organization={Springer}
}

@article{chen2022plot,
  title={Plot: Prompt learning with optimal transport for vision-language models},
  author={Chen, Guangyi and Yao, Weiran and Song, Xiangchen and Li, Xinyue and Rao, Yongming and Zhang, Kun},
  journal={arXiv preprint arXiv:2210.01253},
  year={2022}
}

@article{liang2025comprehensive,
  title={A comprehensive survey on test-time adaptation under distribution shifts},
  author={Liang, Jian and He, Ran and Tan, Tieniu},
  journal=IJCV,
  volume={133},
  number={1},
  pages={31--64},
  year={2025},
  publisher={Springer}
}

@article{gabeff2024wildclip,
  title={WildCLIP: Scene and animal attribute retrieval from camera trap data with domain-adapted vision-language models},
  author={Gabeff, Valentin and Ru{\ss}wurm, Marc and Tuia, Devis and Mathis, Alexander},
  journal=IJCV,
  volume={132},
  number={9},
  pages={3770--3786},
  year={2024},
  publisher={Springer}
}

@article{chi2025learning,
  title={Learning to adapt frozen clip for few-shot test-time domain adaptation},
  author={Chi, Zhixiang and Gu, Li and Liu, Huan and Wang, Ziqiang and Wu, Yanan and Wang, Yang and Plataniotis, Konstantinos N},
  journal={arXiv preprint arXiv:2506.17307},
  year={2025}
}

\end{document}